\documentclass[twoside,11pt]{article}

\usepackage{blindtext}

\usepackage{jmlr2e}

\usepackage{lastpage}

\usepackage{amsmath}
\usepackage{amssymb}
\usepackage{mathtools, nccmath, bbold}

\usepackage{tabularx}
\usepackage{makecell}
\usepackage{multirow}
\usepackage{booktabs}
\usepackage{rotating}
\usepackage{float}

\usepackage{algorithm2e}
\usepackage{algpseudocode}
\algdef{SE}[DOWHILE]{Do}{doWhile}{\algorithmicdo}[1]{\algorithmicwhile\ #1}%

\usepackage{array}

\usepackage{fixltx2e}

\usepackage{stfloats}

\usepackage{url}
\usepackage[T1]{fontenc}

\ShortHeadings{Time Elastic Neural Networks}{pfm}
\firstpageno{1}

\begin{document}
	
\title{Time Elastic Neural Networks}
	
\author{\name Pierre-Fran\c{c}ois Marteau, \email pierre-francois.marteau@univ-ubs.fr \\
	\addr Institut de Recherche en Informatique et Syst\`{e}mes Aléatoires (IRISA)\\ 
	Universit\'{e} Bretagne Sud\\
	Vannes, France.
	}

\editor{My editor}	
	
\maketitle

\begin{abstract}
We introduce and detail an atypical neural network architecture, called time elastic neural network (teNN), for multivariate time series classiffcation. The novelty compared to
classical neural network architecture is that it explicitly incorporates time warping ability that is based on elastic kernel theory, as well as a new way of considering attention. In addition, this architecture is capable of learning a dropout strategy, thus optimizing its own architecture.

The experiment demonstrates that the stochastic gradient descent implemented to train
a teNN is quite effective. While maintaining good accuracy compared to state of the art deep learning approaches, we get a drastic gain in scalability by first reducing the required number of reference time series, i.e. the number of teNN cells required. Secondly, we demonstrate that, during the training process, the teNN succeeds in reducing the number of neurons required within each cell. Finally, we show that the analysis of the activation and attention matrices as well as the reference time series after training provides relevant information to interpret and explain the classification results.
\end{abstract}
	

%
%
%

\section{Introduction}\label{sec:introduction}

This paper presents a neural network architecture for time series classification that explicitly incorporates time warping capability. Behind the design of this architecture, our overall objective is threefold: firstly, we are aiming at improving the accuracy of instance based classification approaches that shows quite good performances as far as enough training data is available. Secondly we seek to reduce the computation time inherent to these methods to improve their scalability. In practice, we seek to find an acceptable balance between these first two criteria. And finally, we seek to enhance the explainability of the decision provided by this kind of neural architecture.

The approach we develop in this study is rooted into the theory of kernels \citep{Schoenberg38}, which are essentially similarity measures to which an inner product corresponds in the so-called Reproducing Kernel Hilbert Space.

More precisely, the proposed architecture, called time elastic Neural Network (teNN) is derived directly from the Dynamic Time Warping Kernel (KDTW) proposed in \citep{MarteauGibet2014} and its novelty, compared to classical neural networks is the following.
\begin{enumerate}
	\item The full network architecture is an assembly of competitive sub-networks called teNN cells. Each teNN cell is associated with three main components: i) an abstract time series, called a reference, ii) an activation matrix and iii) an attention matrix.
	\item Within a cell, the output of any elementary neuron is the sum of its inputs (at most three) multiplied by the local kernel that evaluates the pairwise matching of time series samples (thus, the samples of the input time series are compared to that of the reference time series).
	\item The inverse of the bandwidth of the local kernel is a parameter ($\nu$) that is learned during training. A large value means high local attention, while a small value means a low attention, an area where we do not care any sample comparison. All attention parameters (inverse of the bandwidth values) are gathered within a teNN cell into an attention matrix.
	\item each elementary neuron is associated to an activation weight that is learned during the training. Therefore, inactivated neurons after training can be dropped to simplify the neural architecture. In a way, the network is able to optimize its own architecture. All activation weights in a teNN cell are gathered into an activation matrix.  
	\item Finally, the samples of the reference time series are also learned during training.
\end{enumerate}

While we expect good accuracy, we also expect a drastic gain in scalability by first reducing the number of references, i.e. the number of teNN cells required. Secondly, we hope to reduce the number of neurons required in a cell. Finally, we believe that the analysis of activation and attention matrices as well as references after learning will provide relevant information for interpreting and explaining classification results.

The remaining par of the article is organized as follows. In section 2, we present a brief history of relevant works in the domain of time elastic matching, going from early definition of elastic distances to elastic kernel. Then, as the proposed architecture is essentially inspired from the implementation of KDTW, we detail the way this kernel has been constructed, some of its properties and its implementation in section 3. Section 4 is dedicated to the presentation of the proposed neural architecture. The differentiation of the teNN cells is detailed in section \ref{sec:teNN_differentiation}. A stochastic gradient descent is proposed to minimize a categorical cross entropy loss covering the entire teNN architecture. Section \ref{sec:validation} shows some results obtained on synthetic and real datasets. We confront here our expectations to the experimental reality. In section \ref{sec:experimentation} we compare teNN accuracy to the state of the art in multivariate time series classification before concluding this study.

\section{A brief history of time elastic matching and the root of time elastic neural network}
\begin{figure*}[!ht]
	\centering
	\includegraphics[scale=.52]{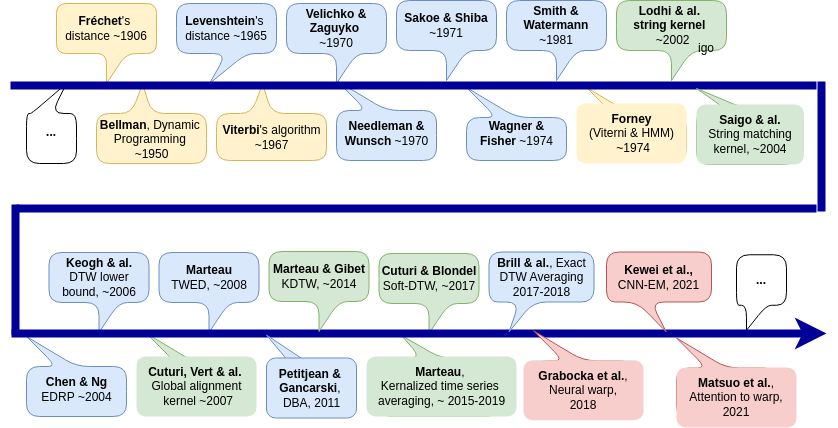}
	\caption{A non exhaustive history of time elastic matching for time series comparison. The founding work is presented in orange, the work on elastic distances in blue and the work on elastic kernel in green.}
	\label{fig:history}
\end{figure*}

The following survey on time elastic matching, that spans more than a century, as depicted in Fig.\ref{fig:history}, is indeed not exhaustive. We mostly focus on the works or results which seem enlightening to us in the context of the study presented in this article. 

The concept of temporal elastic matching between two curves, $x$ and $y$, was historically introduced by Maurice Fr\'{e}chet~\citep{Frechet1906} in the form of an eponymous distance. According to its mathematical formulation given in Eq.\ref{eq:Frechet}, this pairwise distance is defined as an optimization problem in the space of monotonically increasing (temporal) functions. Formally, Fr\'{e}chet defined the pairwise distance measure $F$ between two time series $x$ and $y$ as:

\begin{eqnarray}
	\label{eq:Frechet}
	&F(x,y) = \underset{\alpha, \beta}{Inf}\,\,\underset{t \in [0,1]}{Max} \,\,  \Bigg \{d \Big ( x(\alpha(t)), \, y(\beta(t)) \Big ) \Bigg \}\\
	&\forall t\texttt{, }\forall \delta_t>0\texttt{, }\alpha(t) \le \alpha(t+\delta_t)\texttt{ and }\beta(t) \le \beta(t+\delta_t)\nonumber
\end{eqnarray}

with $\alpha(t)$ and $\beta(t)$ two monotonic increasing temporal functions on which the optimization applies.

Hence, according to Fr\'{e}chet's metaphoric illustration, the distance between two (3D) trajectories followed by a man accompanied with his dog is the minimum length of a leash required to connect the dog and its owner as they walk freely, but without going backward, along their respective paths from one endpoint to the other.

To our knowledge, this is the first time that the temporal way in which curves are traveled has been explicitly taken into account into the design of a distance between two time series. Elastic time matching was born.
 
Obviously, this formal definition was impossible to calculate efficiently in 1906, and it was not until the 1960s and 1970s, with the advent of the early computers, that we saw the first implementations of distance or similarity functions sharing with Fréchet's distance  this concept of temporal elasticity. Earlier and latest implementations of such time elastic distances are based on the optimality principle developed by Bellman \citep{Bellman1957}. Once the so-called \textit{Dynamic Programming} (DP) algorithm has been proposed by Bellman to solve complex (exponential) resource allocation problems in polynomial time, applications to time elastic matching spread rapidly across the computer science community. The Viterbi algorithm \citep{Viterbi:1967}, used in Hidden Markov Model \citep{Forney1973} to align with some elasticity a sequence of observable with a sequence of hidden states, paved the way to the use of DP in the scope of elastic matching.
Subsequently to the Viterbi algorithm, Dynamic time Warping (DTW) was proposed \citep{VelichkoZagoruyko1970, SakoeChiba1971} in the context of speech recognition and then widely generalized to numerous application areas. About at the same time, Needlemann and Wunsch \citep{NeedlemanWunsch1970} proposed an eponym algorithm to evaluate, using DP, the global maximal alignment of two strings. It has been widely used in bio-informatics, to align protein or nucleotide sequences.  
One can also mention the Levenshtein distance (also called edit distance) for string comparison \citep{Levenshtein1966} that was originally proposed in the 1960s found ten years later a DP implementation \citep{WagnerFischer1974} solving the pairwise distance evaluation in $O(n^2)$ complexity that greatly generalized its use, in particular as a spell checker and guesser. Its adaptation to the field of bioinformatics was proposed by Smith and Watermann \citep{SmithWaterman1981}. Subsequently, other proposals seeking to satisfy the triangular inequality unsatisfied by DTW emerged such as the Edit Distance with Real Penalty (EDR)\citep{Chen2004} and the Time Warp Edit Distance (TWED) \citep{Marteau2008}. 

Meanwhile, the advent of support vector machines (SVM) in the early 1990s shed light on the theory of kernels~\citep{Schoenberg38}, opening a new path toward the development of time elastic kernel. First string kernels have been proposed \citep{Lodhi:2002, Saigo2004} then elastic kernels for time series matching were designed, in particular the Global Alignment Kernel (GAK), \citep{Cuturi2007} and the KDTW kernel \citep{MarteauGibet2014}. 

\begin{figure}[]
	\centering
	\includegraphics[scale=1]{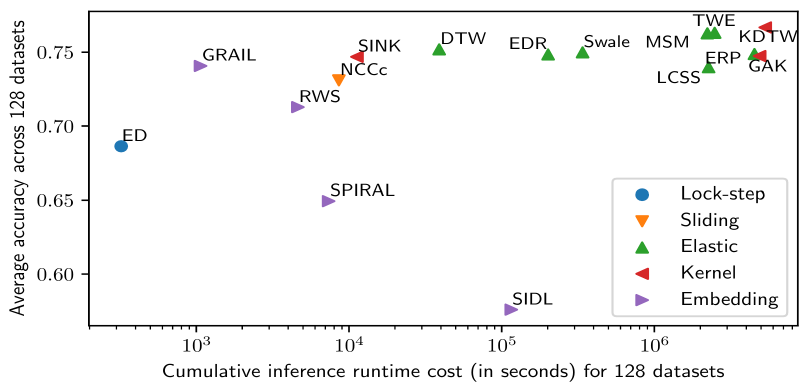} \\
	\caption{Ranking of elastic measures according to Paparrizos et al. study (Figure from \citep{Paparrizos2020}). Distance and kernel measures are evaluated on 128 datasets from the UCR archive.}
	\label{fig:rankingSOT}
\end{figure}
	
Some of these elastic distances have been intensively evaluated on numerous classification tasks such as in~\citep{bagnall2016great} and ~\citep{middlehurst2024bake}. The study by Paparrizos et al. (\citep{Paparrizos2020}) specifically focusing on distance and kernels evaluation on 1NN classification tasks exploiting a set of 128 dataset from the UCR repository \citep{UCRArchive} is likely to be the most exhaustive. Fig.\ref{fig:rankingSOT} showing that kernels compete advantageously (KDTW, GAK) in terms of accuracy,  but at the cost of computational efficiency.

Elastic dissimilarity or similarity functions have more recently been the subject of complementary research to solve the problem of estimating the time elastic mean of a set of time series. On the similarity track essentially based on the DTW we can highlight the DBA algorithm \citep{Petitjean2011} and a proposal for exact calculation of a DTW average \citep{Brill2018}. Regarding the kernel track, we mainly identify Soft-DTW \citep{Cuturi2017} and a kernelized version of DBA called TEKA (Time Elastic Kernelized Averaging) \citep{Marteau2019}.

Algorithmically speaking, DP allows to reduce computational complexity of all these time elastic measures to a polynomial function of the time series lengths, in general of degree 2. 

To conclude this short survey, successful attempts to integrate warping capability into deep neural networks architectures have been develops. In \citep{Grabocka2018} the so-called NeuralWarp is composed with a bi-directional recurrent neural network followed by 4 layers of fully connected neurons used to learned a parametric warping function. Addressing the issue from the angle of convolution kernel, Kewei et al. \citep{Kewei2021} proposed to directly incorporate a warping binary matrix into convolutional layer and evaluated with great success these new layers into common architectures such as ResNet or Inception networks architectures. Finally, as part of a metric learning paradigm, \citep{Matsuo2021} proposed learning temporal warping patterns via an attention model where attention (weight) forms the allowed warping paths.

The approach we develop below is complementary to these recent attempts to endow neural networks with temporal warping capabilities. Rooted in kernel theory, this approach is theoretically well-founded and proposes a new neural architecture that implements an original model of attention not yet proposed to our knowledge.Finally, it offers a new way of reconciling classification accuracy and computational parsimony, while still ensuring  explainability of the results.

\section{The KDTW kernel}
In the following subsections, we detail the construction of the KDTW kernel, on which we based the development of the teNN architecture.

\subsection{Few definitions}
The following definitions will be used through out the article.

\begin{definition}{Time series:}
\label{def:timeseries}
\begin{enumerate}
	\item A (discrete) time series is considered through out this article as a sequence $x=x(0), x(21),\cdots,x(|x|-1)$ of multidimensional samples $x(i) \in S \cup \{\Lambda\}$, where $|x|$ is the length of $x$ and $\Lambda$ is the null sample element. In general $S \subset \mathbb{R}^d$, with $d \in \mathbb{N}^+$, but it could be also a set of finite discrete symbols for instance. In the remaining part of the article we will consider that $S \subset \mathbb{R}^d$
	\item Let $x_n$, with $0 \le n < |x|$, be the truncated time series obtained from $x$ up to sample $n$ ($x_n = x(0)x(1),\cdots, x(n-1)$).
	\item ${}^rx$ will denote the time series obtained when reverting time series $x$, basically ${}^rx(i) = r(|x|-i-1)$, for all $i \in \{0,1,\cdots,|x|-1\}$
	\item Finally, by convention, if $k<0$ or $k\ge |x|$ we consider that $x(k) = \Lambda$ (padding).
\end{enumerate}

Let $\mathcal{U}$ be the set of time series and $\Omega \in \mathcal{U}$ be the null time series (time series of length $0$). We will denote $\mathcal{U}_n = \{x\in \mathcal{U} \texttt{  s.t. }|x| \le n\}$ the set of time series whose size is lower or equal to $n$.
\end{definition}

\begin{definition}{Alignment map:}
	\label{def:alignmentMap}
Let $\pi$ be an ordered alignment map between two finite non empty sequences of successive integers of length $n$ and $m$ respectively. Basically $\pi$ is a finite sequence of pairs of integers $\pi(l)=(i_l, j_l)$ for $l \in \{0,..., |\pi|-1\}$, satisfying the following conditions
\begin{enumerate}
	\item $0 \le i_l, \forall l \in {0, .., |\pi|-1}$
	\item $i_{l} \le i_{l-1}+1, \forall l \in {1, .., |\pi|-1}$ 
	\item $j_{l} \le j_{l-1}+1, \forall l \in {1, .., |\pi|-1}$
	\item $i_{l-1} < i_l$ or $j_{l-1} < j_l, \forall l \in \{1, .., |\pi|-1\}$
\end{enumerate}
$\pi_1(l)=i_l$ and $\pi_2(l)=j_l$ are the two coordinate access functions for the $l^{th}$ pair of mapped integers so that $\pi(l)=(\pi_1(l), \pi_2(l))$. 

We refer to the alignment map that is symmetrical to $\pi$ as $\tilde{\pi}$, namely, $\tilde{\pi}_1 = \pi_2$ and $\tilde{\pi}_2 =\pi_1$.

For all $n \ge 1$ and $m \ge 1$, let $\Pi_{n,m}$ be the set of alignment maps $\pi$ such that the two sets of mapped integers by $\pi$ are $\{1 \cdots n\} \times \{1 \cdots m\}$. 
\end{definition}

\begin{figure}[]
	\centering
	\includegraphics[scale=0.5]{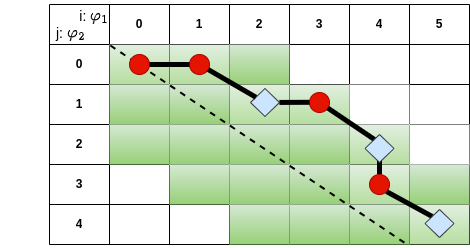}
	\caption{Example of an alignment path corresponding to the alignment map $(0,0)(0,1)(1,2)(1,3)(2,4)(3,4)(4,5)$. The white squares correspond to substitution or match operations and black circles to either deletion or insertion operations.}
	\label{fig:piMathPath}
\end{figure}

Fig.\ref{fig:piMathPath} gives an example of an alignment map that corresponds to an alignment path traversing the $n \times m$ grid while satisfying the conditions specified in Definition~\ref{def:alignmentMap}.

\subsection{Dynamic Time Warping}

Using the previous definition of an alignment map and corresponding path, the DTW measure between two time series $x$ and $y$ is straightforwardly defined as:
\begin{equation}
\label{eq:dtw}
\delta_{dtw}(x,y) =\underset{\pi \in \Pi_{|x|,|y|}}{min} \sum_{i=1}^{|\pi|}(x(\pi_1(i))-y(\pi_2(i)))^2
\end{equation}

However, solving directly the optimization problem that defines DTW (Eq.\ref{eq:dtw}) is difficult since the number of available paths corresponding to a valid alignment map in a $n\times m$ grid is known to be a Delannoy's number, $D(n,m)$~\citep{Banderier2005}. When $n$ and $m$ are in the same order, this number asymptotically increases as $D(n) = {\frac {c\,\alpha ^{n}}{\sqrt {n}}}\,(1+O(n^{-1}))$ where $n\times n$ is the size of the square grid, $\alpha \approx 5,828$ and $c=\approx 0,5727$. Hence, the above DTW optimization problem consists in searching an optimal path in a set of paths whose cardinal increases exponentially with the length of the compared time series.

This is where Bellman's optimality principle \citep{Bellman1957} and the dynamic programming paradigm come into play, allowing to derive in a recursive way the optimal alignment path with a quadratic computational time complexity \citep{SakoeChiba1971}.
	
\begin{eqnarray}
\label{Eq:dtwdp}
\delta_{dtw}(x(n),y(m))&= &(x(n),y(m))^2\nonumber \\
&+&Min
\left\{
\begin{array}{ll}
	\delta_{dtw}(x(n-1),y(m))\\
	\delta_{dtw}(x(n-1),y(m-1))\\
	\delta_{dtw}(x(n),y(m-1))\\
\end{array}
\right.
\end{eqnarray}

To further reduce the time complexity, Sakoe and Chiba proposed to limit the search space to a symmetric corridor disposed around the main diagonal of the grid. The green cells of the grid presented in Fig. \ref{fig:piMathPath} illustrates this kind of corridor.

\subsection{Kernelization of DTW}

Since it has been shown \citep{Lei2007,Marteau2014} that it is not possible to derive directly definite (positive or negative) kernels from the elastic distances mentioned previously, including DTW, the kernelization of DTW has attracted some attention during the last decade. Global alignment Kernel (GAK) \citep{CuturiVert2007}, Kernalized DTW (KDTW) \citep{MarteauGibet2014} and soft-DTW\citep{Cuturi2017}  are some of the most prominent approaches in this area. As our proposal for a time elastic neural network is directly derived from KDTW, we detail below how this positive definite kernel has been originally elaborated.

\begin{definition}{$\pi$-embeding: }
	\label{piEmbeddings}
	For all $n>1$ and all $\pi \in \Pi_{n,n}$, we introduce two projections for time series , basically two vectorized representations,  $\varphi_{\pi_1}: \mathbb{U}_n \rightarrow U_{2n,\pi}$ and $\varphi_{\pi_2}: \mathbb{U}_n \rightarrow U_{2n,\pi}$. These projections are uniquely induced by the alignment map $\pi$. 
	Here $U_{2n,\pi} \subset (\mathbb{R}^d \cup \{\Lambda\})^{2n}$ can be considered as a subset of times series whose lengths are at most $2n$.
	
	Note that the maximal length of an alignment map, as defined by Def.\ref{def:alignmentMap}, allowing to align two time series of length $n$ is $2n$.
	
	Then, the principle of constructing these two projections is simple. Given any alignment map $\pi \in \mathcal{M}_n$ and any time series $x$, we traverse $\pi$ step by step from $l = 1$ to $l = |\pi|$, while applying the following rules:
	\begin{enumerate}
		\item $\varphi_{\pi_1}(x)(0) = \varphi_{\pi_2}(x)(0) = 0$
		\item if both indexes $\pi_1(l)$ and $\pi_2(l)$ increase, then, we set $\varphi_{\pi_1}(x)(l) = x(\pi_1(l))$  and  $\varphi_{\pi_2}(x)(l) = x(\pi_2(l))$,
		\item if only index $\pi_1(l)$ increases, then we set in $\varphi_{\pi_1}(x)(l) = x(\pi_1(l))$ and $\varphi_{\pi_2}(x)(l) = \varphi_{\pi_2}(x)(l-1)$, 
		\item if only index $\pi_2(l)$ increases, then we set $\varphi_{\pi_1}(x)(l) = \varphi_{\pi_1}(x)(l-1)$ and $\varphi_{\pi_2}(x) = x(\pi_2(l))$,  
		\item when we reach the end of $\pi$, if the lengths  of $\varphi_{\pi_1}(x)$ (respectively  $\varphi_{\pi_2}(x)$) is shorter than $2n$, then we insert $\Lambda$ into the remaining slots.\\
	\end{enumerate}

	If we consider the example given in Fig.\ref{fig:piMathPath}, for any time series $x \in U_6$ corresponds two projections in $U_{12} \subset (\mathbb{R}^d \cup \{\Lambda\})^{2\times 6}$ given the alignment map. These projections are:
	
	\begin{align}
		\varphi_{\pi_1}(x) = [x(0),x(1),x(2),x(3),x(4),x(4),x(6),\Lambda,\Lambda,\Lambda,\Lambda,\Lambda]\\ 
		\varphi_{\pi_2}(x) = [x(0),x(0),x(1),x(1),x(2),x(3),x(4),\Lambda,\Lambda,\Lambda,\Lambda,\Lambda] \nonumber 
	\end{align}

	Finally, for any $x \in \mathbb{U}_n$ and $\pi \in \Pi_{n,n}$, we denote $\mathcal{P}_{\pi}(x)=\{\varphi_{\pi_1}(x),\varphi_{\pi_2}(x)\}$ the set of projections (or parts) for time series $x$ induced by $\pi$. Note that all these projections are sequences whose lengths are $2n$. 
\end{definition}  

\begin{figure}[]
	\centering
	\includegraphics[scale=0.4]{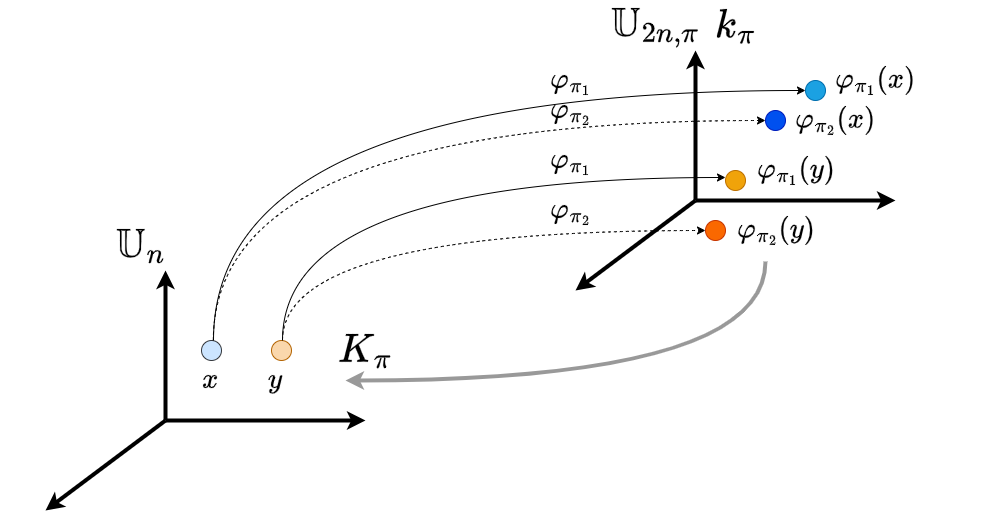}
	\caption{Projections generated by the alignment path $\pi$. To each time series in $\mathbb{U}_n$ corresponds two series (embeddings) in the space $\mathbb{U}_{2n}$. The existence of a kernel in the embedding space allows for the construction of an elastic kernel back into the time series space $\mathbb{U}_n$.}
	\label{fig:embedding}
\end{figure}

\begin{proposition}
	\label{prop:embeddingKernel}
	If kernel $k(.,.)$ is positive definite on $\mathbb{R}^d \cup \{\Lambda\}$ then $\forall n \ge 1$ and $\forall \pi \in \Pi_{n,n}$, then 
	\begin{equation}
		k_\pi(a,b) = \prod_{l=1}^{2n} k(a(l), b(l))
    \end{equation}	
    is a p.d. kernel on $(\mathbb{U}_{2n})$
\end{proposition}

\begin{proposition}
	\label{prop:globalPathKernel}
	If kernel $k(.,.)$ is positive definite on $\mathbb{R}^d \cup \{\Lambda\}$ then $\forall n \ge 1$ and $\forall \pi \in \Pi_{n,n}$, 
	\begin{equation}
		K_{\pi}(x,y)=\sum_{\varphi(x) \in \mathcal{P}_{\pi}(x)} \sum_{\varphi(y) \in \mathcal{P}_{\pi}(y)} \prod_{l=1}^{2n} k(\varphi(x)(l), \varphi(y)(l)) \\
	\end{equation}
	is a p.d. kernel on $(\mathbb{U}_n)$. \\
\end{proposition}

Proof of proposition \ref{prop:globalPathKernel} is a direct consequence of the Haussler's \textit{R-convolution} kernel theorem \citep{Haussler99}. Indeed, since $k(x,y)$ is a p.d. kernel on $(\mathbb{R}^d \cup \{\Lambda\})$, and, considering the sets of parts  $\mathcal{P}_{\pi}(x)$ and  $\mathcal{P}_{\pi}(y)$ associated respectively to the sequences $x$ and $y$, the conditions for the Haussler's \textit{R-convolution} are satisfied.\\

Note that $K_{\pi}(x,y)$ simply rewrites as
\begin{align}
	K_{\pi}(x,y)=\prod_{l=1}^{2n}k(\varphi_{\pi_1}(x)(l), \varphi_{\pi_2}(y)(l)))\nonumber \\
	+\prod_{l=1}^{2n}k(\varphi_{\pi_2}(x)(l), \varphi_{\pi_1}(y)(l)))\nonumber \\
	+\prod_{l=1}^{2n}k(\varphi_{\pi_1}(x)(l), \varphi_{\pi_1}(y)(l))) \label{eq:kpi}\\ 
	+\prod_{l=1}^{2n}k(\varphi_{\pi_2}(x)(l), \varphi_{\pi_2}(y)(l)))\nonumber  
\end{align}

In practice, $k(a,b) = \frac{1}{3}e^{-\nu(a-b)^2}$ is chosen as the local positive kernel defined on $\mathbb{R}^d \cup \{\Lambda\}$, with $\nu \in \mathbb{R}^+$ and considering for instance $(a-\Lambda)^2 = \infty$, $\forall a \in \mathbb{R}^d\setminus\{\Lambda\}$.

Let $\mathcal{C}_{2n} \subset \mathcal{M}_{2n}$ be a subset of paths then the following holds. 

\begin{proposition}
	\label{prop1}
	For any $n>1 \in \mathbb{N}$, any $\pi \in \mathcal{C}_{2n} \subset \mathcal{M}_{2n}$, and any $(x,y) \in (\mathbb{U}_n)^2$,  then the following kernel is positive definite on  $\mathbb{U}_n$
	\begin{equation}
		KDTW(x,y) = \sum_{\pi \in \mathcal{C}_{2n}} K_{\pi}(x,y)
	\end{equation}
\end{proposition}

The three previous propositions are also a consequence of the fact that the set of positive definite kernels is a closed (w.r.t. pointwise convergence) convex cone stable under addition and multiplication. 

The choice for the local kernel $k(a,b) = \frac{1}{3}e^{-\nu(a-b)^2}$ in the embedding space allows for a probabilistic interpretation of $KDTW$. It provides a probability (up to a normalization factor) for the matching of samples $a$ and $b$. As we multiply these local matching probabilities along the alignment path $\pi$, kernel $k_{\pi}$ provides a distribution of probability (up to a normalization factor) on the set of alignment paths (represented by the four alignments parts listed in Eq.\ref{eq:kpi}). Finally, $KDTW(x,y)$ can be understood as the sum of the probabilities of admissible paths that align $x$ and $y$. 
More details about the proofs and interpretation of $KDTW(x,y)$ can be found in \citep{MarteauGibet2014}.

The following equations define a recursive implementation (thanks to the dynamic programming solution) of KDTW that can be evaluated with quadratic complexity ($O(n^2)$) in both time and space. It includes a corridor, $h$, symmetrical to the main diagonal of the alignment grid.

\begin{equation}
	\label{eq:kdtw-r}
	KDTW(x,y) = \mathcal{C}_{h}(x, y) + \tilde{\mathcal{C}}_{h}(x, y)
\end{equation}	

with the two recursive equations starting with $p = |x|$ and $q = |y|$.

\begin{align}
	&\mathcal{C}_{h}(x_p, y_q)= \frac{1}{3} e^{-\nu (x(p) - y(q))^2}\nonumber\\
	& \sum \left\{
	\begin{array}{ll}
		h(p-1,q) \mathcal{C}_{e,h}(x_{p-1},y_q)\hspace{1mm} \\
		h(p-1,q-1) \mathcal{C}_{e,h}(x_{p-1},y_{q-1})\hspace{1mm} \label{eq:kdtw1} \\
		h(p,q-1) \mathcal{C}_{e,h}(x_{p},y_{q-1})\hspace{1mm}  
	\end{array}
	\right.
\end{align}

\begin{align}
	&\tilde{\mathcal{C}}_{h}(x_{p}, y_{q}) = \frac{1}{3} \nonumber\\
	& \sum \left\{
	\begin{array}{ll}
		h(p-1,q) \tilde{\mathcal{C}}_{h}(x_{p-1},y_{q}) (e^{-\nu (x(p) - y(p))^2} + e^{-\nu (x(q) - y(q))^2})\label{eq:kdtw2}\\
		\frac{1}{2}\delta_{pq} h(p-1,q-1) \tilde{\mathcal{C}}_{h}(x_{p-1},y_{q-1}) e^{-\nu (x(p) - y(p))^2} \\
		\frac{1}{2}h(p,q-1) \tilde{\mathcal{C}}_{h}(x_p,y_{q-1}) (e^{-\nu (x(p) - y(p))^2} + e^{-\nu (x(q) - y(q))^2}) 
	\end{array}
	\right.
\end{align}

By construction, the first term in Eq.\ref{eq:kdtw-r} evaluates over all the considered alignment paths the two first products listed in Eq.\ref{eq:kpi}, while the second term in Eq.\ref{eq:kdtw-r} evaluates over all the alignment paths the two last products listed in Eq.\ref{eq:kpi}.

In the two terms of the recursive equations (Eq.\ref{eq:kdtw1} and Eq.\ref{eq:kdtw2}), the $h(p,q)$ function represents a symmetric corridor that can be defined along the main diagonal of the alignment grid.

In the second term (Eq.\ref{eq:kdtw2}) $\delta_{pq}$ is the Kronecker's symbol: 
$\delta(p,q)={\begin{cases}1&{\mbox{if }}p=q\\0&{\mbox{if }}p\neq q\end{cases}}$

\subsection{The forward and backward kernel evaluation matrices}
The recursive equations of KDTW (Eq.\ref{eq:kdtw-r},\ref{eq:kdtw1},\ref{eq:kdtw2})  allow to construct a $|x|\times|y|$ forward matrix $F(x,y)$ whose elements $F(x,y)_{i,j}$ are nothing but the evaluation of the kernel on the prefix time series $x_i$ and $y_j$, namely $KDTW(x_i,y_j)$, i.e. the sum of the probabilities of all the alignment paths that align $x$ and $y$ up to samples $i$ and $j$ respectively, or in other words, the sum of the probabilities of all the the alignment paths connecting cell $(0,0)$ to cell $(i,j)$ of the alignment grid.  

Similarly, we construct the backward alignment matrix, $B(x,y)$, from the reverse time series ${}^rx$ and ${}^ry$. More precisely, $B(x,y)$ is defined such that $B(x,y)_{i,j} = KDTW({}^rx_i,{}^ry_j)$.

\begin{figure}[H]
	\centering
	\begin{tabular}{ccc}
		\includegraphics[scale=.8, angle=0]{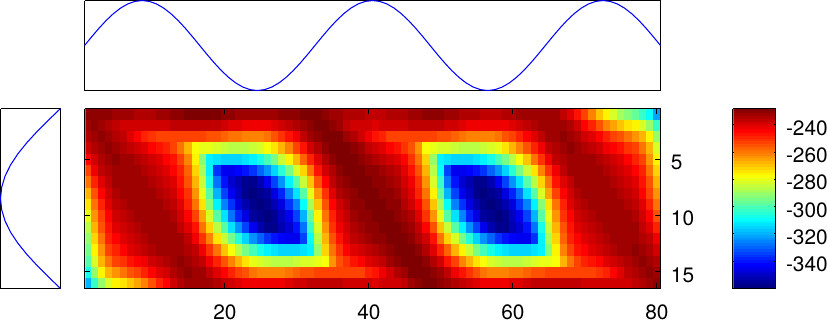} &
	\end{tabular}
	\caption{Forward Backward matrix (logarithmic values) for the alignment of a positive halfwave with a sinus wave. The dark red color represents high probability cells, while dark blue color represents low probability cells.}
	\label{fig:SinTest}
\end{figure}

And the Forward-Backward alignment matrix $FB(x,y)$ is defined as the point-wise multiplication of the forward and backward matrices 

\begin{equation}
	FB(x,y)_{i,j} = F(x,y)_{i,j}\cdot B(x,y)_{|x|-i,|y|-j}
	\label{eq:FB}
\end{equation}

Hence, cell $(i,j)$ of the $FB$ matrix evaluates (up to a constant) the sum of the probabilities of all the existing alignment paths between $x$ and $y$ that traverse cell $(i,j)$.

The $FB$ matrix and its two components $F$ and $B$ provide interpretive information about the alignment process. As an example, Fig. \ref{fig:SinTest} presents the Forward-Backward matrix corresponding to the alignment of a positive half-wave with a sine wave. The three areas of likely alignment paths are clearly identified in dark red color, while the low probability alignment areas are encoded in dark blue color. \\

In section \ref{sec:teNN_differentiation} the $F$ and $B$ matrices will be used to differentiate the KDTW kernel, whose evaluation is provided by the core cell of the teNN architecture detailed in the following subsections. 

\section{Time Elastic Neural Nets }
\label{sec:teNN_architecture}

As shown in Fig.\ref{fig:rankingSOT} and in Table \ref{exp:results} of the experimentation section, k-NN classifiers based on elastic distances or kernels are fairly good classification models if sufficient training data is available. However, finding neighbors in large, multi-dimensional datasets scales very poorly (especially when distance or kernel evaluation is obtain with a quadratic complexity). In addition, decision results are difficult to explain or interpret, as the local geometry of the multi-dimensional embedding space is often highly non-linear and difficult to visualize. 

In this context, the motivation behind the development of the time-elastic neural network (teNN) is the improvement of k-NN classifiers based on the KDTW measure, an improvement that can be quantified in terms of efficiency, accuracy and interpretability of results.

\subsection{KDTW as a network of cells}

\begin{figure}[]
	\centering 
	\includegraphics[scale=.45]{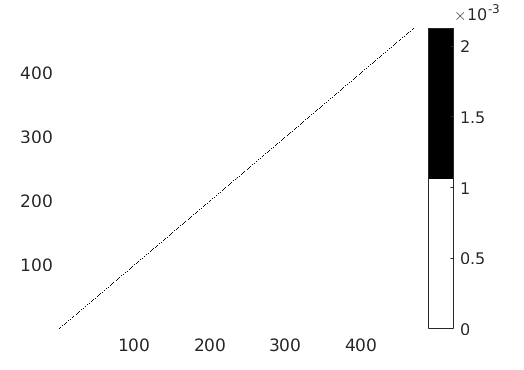} 
	\includegraphics[scale=.45]{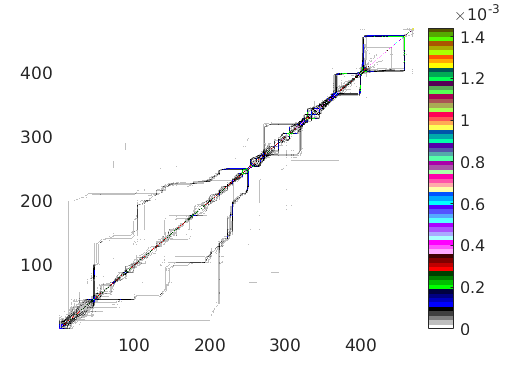} \\
	\caption{UCR Beef dataset: left Sakoe-Chiba 'optimal' corridor, right all the best DTW alignment paths (symmetrized), (from \citep{Soheily2019}.}
	\label{fig:SparseCorridor}
\end{figure}

\begin{figure}[H]
	\centering
	\includegraphics[scale=0.45]{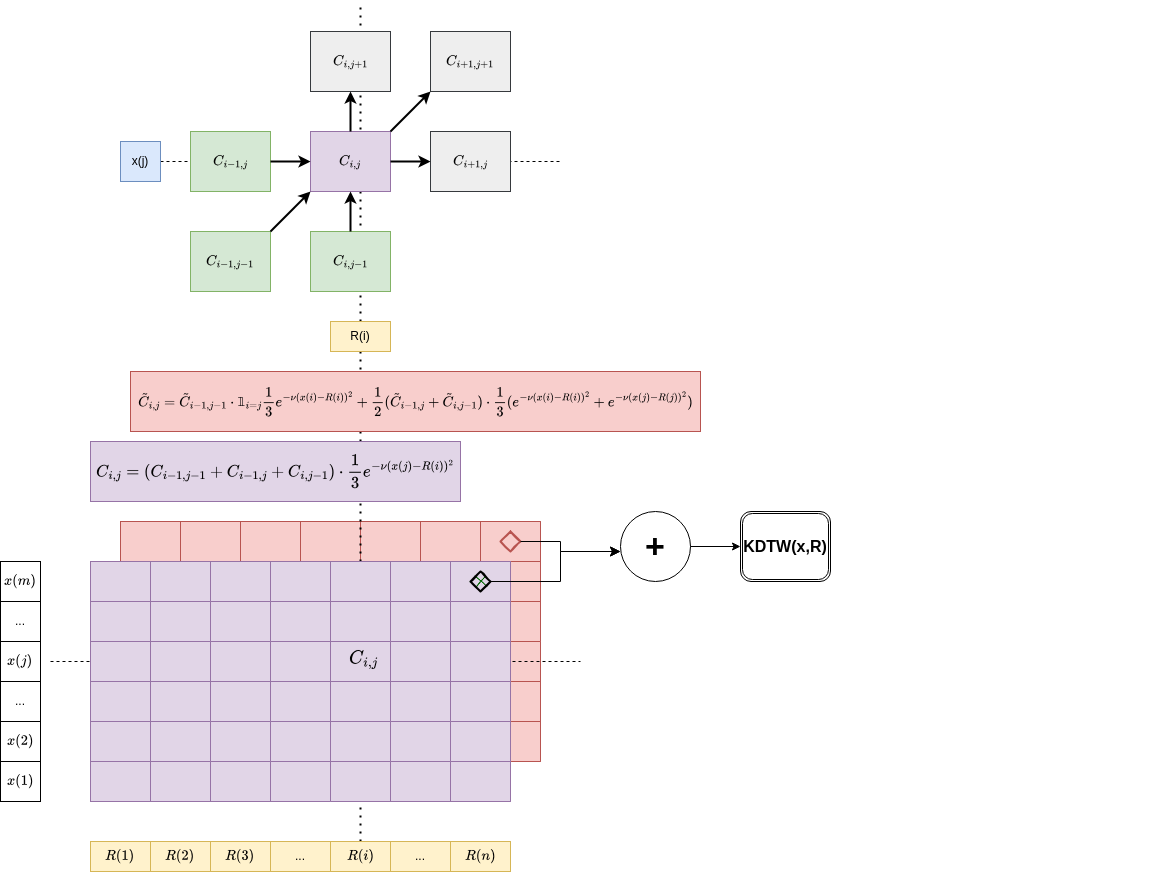}
	\caption{Network architecture dedicated to the computation of the $KDTW$ kernel.}
	\label{fig:kdtwCell}
\end{figure}

Four main considerations guided the design of the teNN architecture.
\begin{enumerate}
	\item The evaluation of KDTW measurement is similar to the diffusion of information in a network of cells. teNN is the result of this analogy taken as far as possible.
	\item To speed up k-NN classifiers, one can significantly reduce the size of the train set by selecting only reference instances that generally characterize class boundaries, e.g. the support vectors involved in support vector machines, for example. However, another path can be followed, which consists of considering the reference instances as model parameters that can be optimized during the training phase.
	\item In the definition of KDTW, the meta-parameter $\nu$ which enters into the calculation of the local kernel $k(a,b) = \frac{1}{3}e^{\nu(a-b)^2} $ does not depend on the position of samples $a$ and $b$ in the matched time series, although it can be adjusted throughout the alignment paths. By making $\nu$ a time-dependent parameter, it may become possible to take into account a varying selectivity of the local alignment kernel capable of encoding an original form of sequential attention.
	\item The corridor defined by the $h(.,.)$ function has a fixed and dense shape (usually it is defined as a rectangle symmetrically adapted along the main diagonal of the alignment grid). As shown in Fig.\ref{fig:SparseCorridor}, the best alignment paths may be confined inside a potentially sparse corridor of any shape. Somehow, the teNN architecture is aiming at learning the shape of the corridor that best balanced the accuracy and computational cost requirements. 
\end{enumerate}

The computation of $KDTW(x,r)$ between a reference time series, $R$, and time series $x$ consists in progressively evaluating the cells of two grids, $[C_{i,j}]$ (Eq.\ref{eq:kdtw1}) and $[\tilde{C}_{i,j}]$ (Eq.\ref{eq:kdtw2}), according to a process that corresponds to the propagation (from left to right) of partial information into a well defined network architecture. This architecture is depicted in Fig.\ref{fig:kdtwCell}. Namely each cell $(i,j)$ of these two grids are systematically connected to three previous cells $(i-1,j)$, $(i-1, j-1)$ and $(i,j-1)$, except for the initial cells $(0,j)$ and $(i,0)$, $i,j \in \{0,1,\cdots\}$. The summation of the upper right cells of the two grids gives the final result, i.e. $KDTW(x,r)$.

\subsection{teNN elementary layer}

\begin{figure}[]
	\centering
	\includegraphics[scale=0.33]{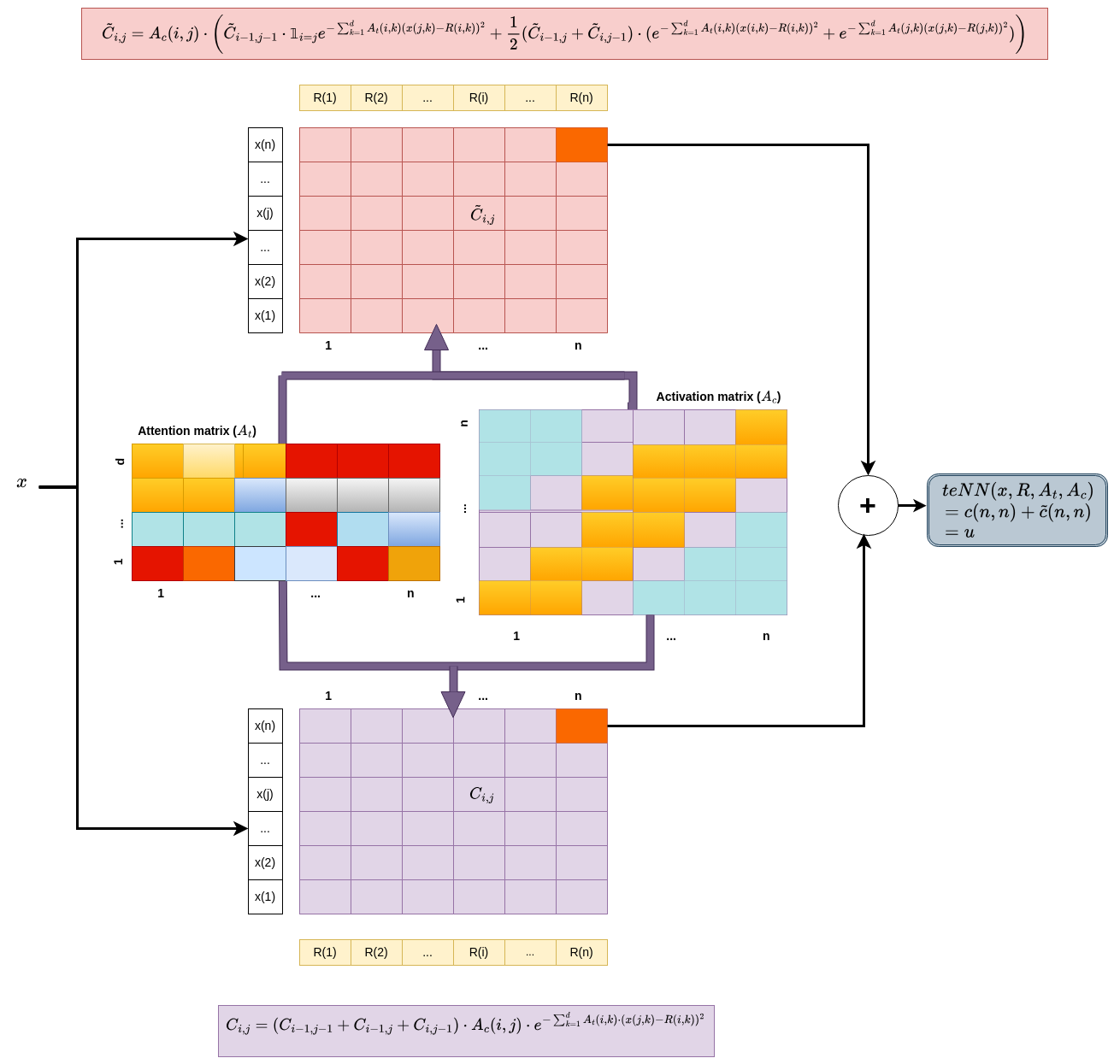}
	\caption{teNN network layer.}
	\label{fig:teNNCells}
\end{figure}

To build a single teNN layer (or component) presented in Fig.\ref{fig:teNNCells}, we are making three major changes to the previous KDTW network architecture. 

First, each layer is composed of a reference matrix $[R(i,k)]$ ($i \in \{0,\cdots, n-1\}$ and $k \in \{0,\cdots, d-1\}$), with $R(i,k)\in \mathbb{R}$, which is trained during the learning phase to best represent a time series subset. More precisely, once trained, $R$ will correspond to a virtual time series that somehow maximizes the sum of pairwise similarity between itself and each of the time series of the considered subset, given the other parameters of the network introduced below. 

Second, to each pair of cells ($C(i,j)$ and $\tilde{C}(i,j)$) we add an activation weight. All activation weights are gathered into an activation matrix, $Ac$, whose elements $Ac(i,j) \in [0;1]$, $\forall i,j=\{0,\cdots,n-1\}$, are aiming at quantifying the activation of the cells at position $(i,j)$. If $Ac(i,j) \rightarrow 0$, then the cell is inactive and the probability for an alignment path to traverse it tends towards $0$.  Conversely, when $Ac(i,j) \rightarrow 1$, the cell is activated. During the training phase, this activation matrix highlights the likely alignment paths and consequently reduces the size of the alignment search space. Basically, it will extract the optimized shape of the alignment corridor.

The last main change is the attention matrix $[A_{t}(i,k)]$ ($i \in \{0,\cdots, n-1\}$ and $k \in \{0,\cdots, d-1\}$). 
The elements of this matrix are positive or null and correspond to the parameter $\nu$ that appears in the local alignment kernel used in KDTW. 
\begin{equation}
	\kappa(R(i),x(j)) = e^{-\sum_{k=1}^d A_{t}(R(i)-x(j))^2}\label{eq:local_kernel}
\end{equation}
When $A_{t}(i,k) = 0$, then the local differences at time stamp $i$ and dimension $k$, $(R(i,k)-x(j,k))^2$, no longer play any role. Conversely, when $A_{t}(i,k)$ takes high values, then the local alignment kernel becomes very selective at time stamp $i$ and dimension $k$. Consequently, visualizing the contents of these matrices enables us to identify the spatio-temporal locations of highly selective patterns, i.e. the area of the reference time series where the network is particularly attentive, and the “don't care” areas where it is mostly inattentive.

In addition, although KDTW allows the management of time series of variable length, we adopt for sake of simplification an architecture able to process time series of length at most $n$. If time series shorter than $n$ are involved, a padding with $\Lambda$ samples is used (in practice, we set $\Lambda = 0$).

\subsubsection{Managing several references per category}

\begin{figure}[H]
	\centering
	\includegraphics[scale=0.7]{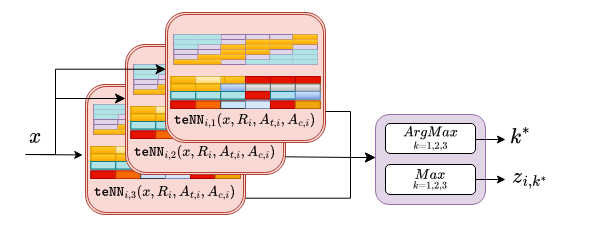}
	\caption{When several references are used to represent a category, the layer whose output is maximum (given an input $x$) is selected.}
	\label{fig:bestRef}
\end{figure}

For some applications, it may be necessary to deal with heterogeneous categories, that is to say containing separated clusters of dissimilar time series. In such situation, the proposed architecture can manage several references in parallel for these heterogeneous categories as depicted in Fig. \ref{fig:bestRef}. When an input $x$ is submitted to a set of elementary layers attached to the same category, the best layer is selected, that is to say the one providing the highest similarity measure in output. This selection is used during training (only the parameters of the best reference will be updated) and during exploitation.

\subsection{teNN full network}

\begin{figure}[]
	\centering
	\includegraphics[scale=0.7]{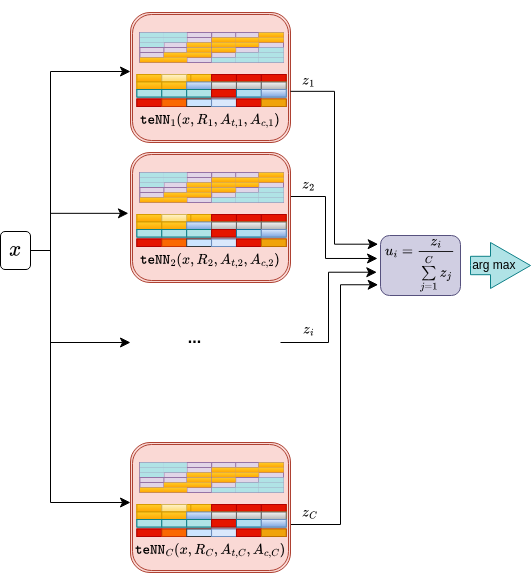}
	\caption{teNN complete network architecture.}
	\label{fig:teNN}
\end{figure}

The complete teNN architecture is depicted in Fig.\ref{fig:teNN}. This is a parallelization of teNN elementary layers. Each reference $R_i$ is associated to a category $y_i$. We can consider either a single reference per category or several, depending on the classification task to be processed. The number of references per category is a meta parameter of the model. The expectation will be that the size of the set of $R_i$ (the number of teNN layers) will be much smaller than the training set. 

As each teNN$_i$ layer output provides, up to a normalization factor, an estimate of the sum of the probabilities of all alignment paths between the tested time series and the reference $R_i$, a final normalization function acting as a softmax layer is preferred over a pure softmax layer. The latter could produce numerical instability due to the exponentiation of non-normalized KDTW values.

The proposed normalizing function, $f$, is simply with $z^t = [z_1,z_2,\cdots,z_C]$:

\begin{equation}
	o_i = f(z)_i = \frac{z_i}{\sum\limits_{j=1}^C z_j}
	\label{eq:final_nomalization}
\end{equation}

\section{teNN differentiation}
\label{sec:teNN_differentiation}
To implement a training procedure for the teNN architecture based on stochastic gradient descent, we need to differentiate the cells of this architecture according to the parameters $(R,A_c, A_{t})$. Below we proceed step by step to provide these required derivatives.\\

\subsection{teNN cell derivatives}
Let begin with a single teNN cell. 
Recall that 
\begin{equation}
	\label{eq:teNN_out}
	\texttt{teNN}(x,R,A_{t}, A_c) = C_{n,n} + \tilde{C}_{n,n} = u
\end{equation}

Let $k(i,j) = \frac{1}{3}\cdot A_c(i,j)\cdot e^{-\sum\limits_{k=0}^{d-1}A_{t}(i,k)(R(i,k)-x(j,k))^2)}$ be the local kernel evaluation at cell $(i,j)$ of the alignment grid ($i,j \in \{0,1,\cdots, n-1\}$, $k \in \{0,1,\cdots, d-1\}$)

\begin{figure}[h!]
	\centering
	\includegraphics[scale=0.7]{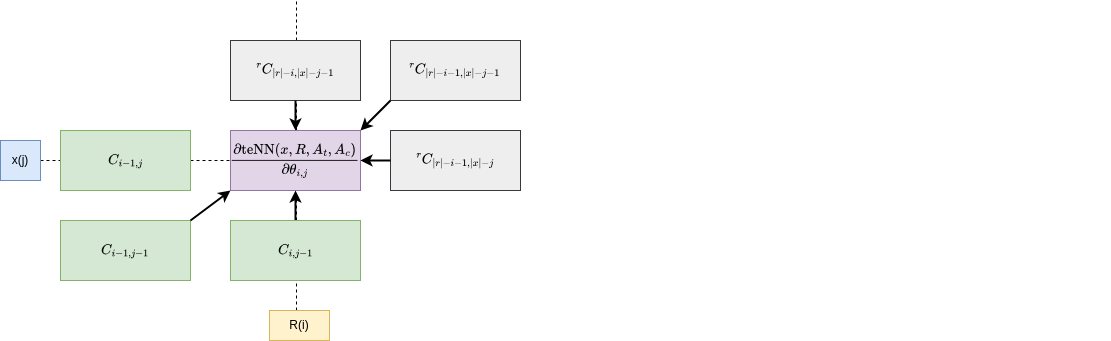}
	\caption{teNN Differentiating cell $C_{i,j}$ (similarly $\tilde{C}_{i,j}$) according to a parameter $\theta_{i,j}$ (that only depends on cell $(i,j)$).}
	\label{fig:FB-derivative}
\end{figure}

\subsubsection{$C_{n,n}$ term differentiation}
Let us first consider the derivative of $C_{n,n}$ according to the three types of parameters that come into play in the teNN cell. By decomposing the derivative process around position $(i,j)$ we get that
\begin{align}
	\frac{\partial C(n,n)}{\partial R(i,k)} = \sum_j\left(\frac{\partial C(n,n)}{\partial k(i,j)} \cdot\frac{\partial k(i,j)}{\partial R(i,k)}\right) \label{eq:dr}\\
	\frac{\partial C(n,n)}{\partial A_{t}(i,k)} = \sum_j\left(\frac{\partial C(n,n)}{\partial k(i,j)} \cdot\frac{\partial k(i,j)}{\partial A_{t}(i,k)}\right) \label{eq:deriv_cnn}\\
	\frac{\partial C(n,n)}{\partial A_c(i,j)} = \frac{\partial C(n,n)}{\partial k(i,j)} \cdot\frac{\partial k(i,j)}{\partial A_c(i,j)} 	
\end{align}

using the derivatives of the local kernel we have

\begin{align}
	\frac{\partial k(i,j)}{\partial r(i,k)} = \frac{2}{3}\cdot A_c(i,j)\cdot A_{t}(i,k)(R(i,k)-x(j,k))\cdot e^{-\sum\limits_{k=0}^{d-1}A_{t}(i,k)(R(i,k)-x(j,k))^2)}\\
	\frac{\partial k(i,j)}{\partial A_{t}(i,k)} = \frac{1}{3}\cdot A_c(i,j)\cdot (R(i,k)-x(j,k))\cdot e^{-\sum\limits_{k=0}^{d-1}A_{t}(i,k)(R(i,k)-x(j,k))^2)}\\
	\frac{\partial k(i,j)}{\partial A_c(i,j)} = \frac{1}{3}\cdot e^{-\sum\limits_{k=0}^{d-1}A_{t}(i,k)(R(i,k)-x(j,k))^2)}
	\label{eq:deriv0}	
\end{align}

For the term $\frac{\partial C_{n,n}}{\partial k(i,j)}$, we make use of the forward matrix $[C_{i,j}]$ and its backward counterpart $[{}^rC_{i,j}]$.

As $k(i,j)$ only occurs in the evaluation of cell  $C_{i,j}$, Fig.\ref{fig:FB-derivative} shows how forward and backward matrices intervene into the determination of the term $\frac{\partial C_{n,n}}{\partial k(i,j)}$ by looking around cell $C_{i,j}$. Basically, we only have to consider all the paths that traverse cell $Ci,j$ to derive $\frac{\partial C_{n,n}}{\partial k(i,j)}$.

According to Fig.\ref{fig:FB-derivative}  the sum $\mathcal{S}_{i,j}$ of all the probabilities of the alignment paths that traverse cell $C_{i,j}$ is given by

\begin{align}
\mathcal{S}_{i,j} = &(C_{i-1,j-1}+C_{i-1,j}+C_{i,j-1}) \cdot k(i,j) \cdot\nonumber\\ 
&({}^rC_{|r|- i-1,|x|- j-1}+{}^rC_{|r|-i-1,|x|-j}+{}^rC_{|r|-i,|x|-j-1})
\end{align}
Hence, we get

\begin{align}
	&\frac{\partial C_{n,n}}{\partial k(i,j)} = \frac{\partial \mathcal{S}_{i,j}}{\partial k(i,j)}\nonumber\\
	&\texttt{   } = (C_{i-1,j-1}+C_{i-1,j}+C_{i,j-1}) \cdot ({}^rC_{|r|- i-1,|x|- j-1}+{}^rC_{|r|-i-1,|x|-j}+{}^rC_{|r|-i,|x|-j-1})
\end{align}

\subsubsection{$\tilde{C}_{n,n}$ term differentiation}

The procedure to obtain the derivatives of the term $\tilde{C}_{n,n}$ is similar to that used for the term ${C}_{n,n}$. We introduce first the function $f(i,j) = \frac{1}{2}(k(i,i)+k(j,j))$ and decompose the partial differentiation as follows

\begin{align}
	\frac{\partial \tilde{C}(n,n)}{\partial R(i,k)} = \sum_j\left(\frac{\partial \tilde{C}(n,n)}{\partial f(i,j)} \cdot\frac{\partial f(i,j)}{\partial R(i,k)}\right) \label{eq:dr_tilde}\\
	\frac{\partial \tilde{C}(n,n)}{\partial A_{t}(i,k)} = \sum_j\left(\frac{\partial \tilde{C}(n,n)}{\partial f(i,j)} \cdot\frac{\partial f(i,j)}{\partial A_{t}(i,k)}\right) \\
	\frac{\partial \tilde{C}(n,n)}{\partial A_c(i,j)} = \frac{\partial \tilde{C}(n,n)}{\partial f(i,j)} \cdot\frac{\partial f(i,j)}{\partial A_c(i,j)} 
	\label{eq:deriv_cnn_tilde}	
\end{align}

Two distinct cases appear for the derivatives of the term $f(i,j)$, depending on whether $i=j$ or not. The formulation below compiles the two possibilities

\begin{align}
	\frac{\partial f(i,j)}{\partial R(i,k)} = \frac{1+\mathbb{1}_{i=j}}{6}\cdot A_c(i,j)\cdot A_{t}(i,k)(R(i,k)-x(i,k))\cdot e^{-\sum\limits_{k=0}^{d-1}A_{t}(i,k)(R(i,k)-x(i,k))^2)}\\
	\frac{\partial f(i,j)}{\partial A_{t}(i,k)} = \frac{1+\mathbb{1}_{i=j}}{6}\cdot A_c(i,j)\cdot (R(i,k)-x(j,k))\cdot e^{-\sum\limits_{k=0}^{d-1}A_{t}(i,k)(R(i,k)-x(j,k))^2)}\\
	\frac{\partial f(i,j)}{\partial A_c(i,j)} = \frac{1}{6}\cdot (e^{-\sum\limits_{k=0}^{d-1}A_{t}(i,k)(R(i,k)-x(i,k))^2} +e^{-\sum\limits_{k=0}^{d-1}A_{t}(j,k)(R(j,k)-x(j,k))^2})
	\label{eq:deriv_tcnn}	
\end{align}

And finally, the exploitation of the forward and backward matrices leads to

\begin{align}
\frac{\partial \tilde{C}_{n,n}}{\partial f(i,j)} = & (\mathbb{1}_{i=j}\tilde{C}_{i-1,j-1}+\tilde{C}_{i-1,j}+\tilde{C}_{i,j-1}) \cdot\nonumber\\ &({}^r\mathbb{1}_{i=j}\tilde{C}_{|r|- i-1,|x|- j-1}+{}^r\tilde{C}_{|r|-i-1,|x|-j}+{}^r\tilde{C}_{|r|-i,|x|-j-1})
\label{eq:end}
\end{align}

\subsection{Loss function for the full teNN architecture training}

In the absence of knowledge on the data, the Categorical Cross-Entropy (CCE) loss seems to be an acceptable choice for training the teNN architecture, although it is known to be sensitive to imbalance data and outliers. Usually, in most neural networks architecture optimization, CCE is used in conjunction with a \textit{softmax} output layer. For the teNN architecture however, we use it with the normalizing function $o$ (Eq.\ref{eq:final_nomalization}). In addition to the CCE loss, we add two regularization terms to force the search for parsimonious solutions. We use the $L_1$ norm to constrain the matrices $A_c$ and $A_t$ of all teNN cells to be sparse. $\lambda_1 \ge 0$ and $lambda_2 \ge 0$ are two meta parameters that weight the importance of these two regularizing terms in the final loss function.

\begin{equation}
	\mathcal{L}(X,y) = - \sum\limits_{l=1}^{|X|}\sum\limits_{i=1}^C y_{l,i}\ln(o_{l,i}) + \lambda_1\sum\limits_{i=1}^C||A_{t,i}||_{L_1} + \lambda_2\sum\limits_{ki1}^C||A_{c,i}||_{L_1}
	\label{eq:loss}
\end{equation}
where $y_i$ is a one-hot vector and $o_{l,i}$ the normalized output vector of teNN layer $i$ (corresponding to the $i^{th}$ category) when $x_l$ is presented as input.\\

The training of the teNN architecture corresponds therefore to the following optimization problem

\begin{eqnarray}
	\Theta^* = 	&\underset{\Theta}{Min}& \mathcal{L}(o,y)\\
	&s.t. & 0 \le A_{c,i} \le 1\nonumber\\
    && 0 \le A_{t,i}\nonumber\\
    && \texttt{for } i \in \{1,2,\cdots,C\}\nonumber
\end{eqnarray}

$\Theta$ is the set of parameters that are optimized, namely $\{(A_{c,k},A_{t,k}, R_k)\}$ for $k \in \{1,2,\cdots,C\}$.

Differentiating the CEL function according to any parameter $\theta_i$ characterizing any teNN layer $i$ can be expressed as:
\begin{equation}
\frac{\partial \mathcal{L}(X,y)}{\partial \theta} =  \frac{\partial \mathcal{L}(o,y)}{\partial \theta} = \frac{\partial \mathcal{L}(o,y)}{\partial z_i}\frac{\partial z_i}{\partial \theta}
\end{equation}
where $o_i$ is the output of teNN layer $i$.\\

$\frac{\partial o_i}{\partial \theta}$ is obtained through the process described from Eq.\ref{eq:teNN_out} to Eq.\ref{eq:end}.\\

And it is easy to get that 
\begin{equation}
\frac{\partial \mathcal{L}(o_i,y)}{\partial z_i} = \frac{1}{z_i}\left(\frac{z_i}{\sum_{j=1}^C z_j} - y_i\right)
\end{equation}

where $y$ is a one hot vector.\\

Finally, to solve the optimization problem, which is quadratic but not convex, we adopt a classical stochastic gradient descent.

\subsubsection{Algorithmic complexity of the training procedure}
For the previous differentiation procedure, the evaluation of the forward and backward alignment matrices ${C}_{n,n}$ and $\tilde{C}_{n,n}$ requires $4n^2$ local kernel evaluations, whose complexity is in $O(d)$. As $n^2$ derivatives need to be evaluated for the $A_c$ (activation) matrix, $nd$ derivatives for the attention matrix $A_{t}$ and  $nd$ derivatives for the reference time series $R$, we get that the overall complexity requirement to differentiate a teNN cell is about $4\cdot d\cdot n^2 + n^2 + 2n\cdot d$ macro operations, leading to a $O(d\cdot n^2)$ time complexity and a $O(n^2+n\cdot d)$ space complexity. hence, for a full teNN architecture containing $N_r$ cells, the computational time and space complexities are $O(N_r\cdot d\cdot n^2)$.

\subsection{Training procedure for teNN}

\SetKwComment{Comment}{/* }{ */}
\RestyleAlgo{ruled}
\begin{algorithm}
	\scriptsize 
	\caption{Stochastic gradient descent on the categorical cross entropy loss for the teNN architecture (Eq.\ref{eq:loss})}\label{alg:CL}
	\KwData{$X$, $y$, $L$, $dim$ \{(teNN$_k$, $y_k, nc_k, f_k$), $k=1\cdots N_c$\}, $max\_epoch$, $batch_size$, $\nu_0$, $\alpha_0$, $\eta$, $\lambda_1$, $\lambda_2$}
	\KwResult{The trained teNN}
	$nepoch \gets 0$\;
	$nbatch \gets max(1,floor(|X|/batch\_size))$\;
	$C \gets set\_of\_categories(y)$\;
	$n \gets length\_of\_time\_series(X)$\;
	$R_{k}  \gets InitializeWithCentroidR(y_k,X,y, nc_k)$, for $k = 0$ to $N_c$\;
	$A_{t,k} \gets \nu_0$, for $k = 0$ to $N_c$\;
	$A_{c,k} \gets \alpha_0$, for $k = 0$ to $N_c$\;	
	\While{$nepoch < max\_epoch$}{
		$list\_of\_batch \gets Random\_split(X, y, nbatch)$\;
		\For{$nc = 0$ to $nbatch$}{
			$X_b, y_b \gets  list\_of\_batch[nc]$\;
			$Gr[N_c,L,dim] \gets  0;$ $GA_c[N_c,n,n] \gets  0;$ $GA_t[N_c,L,dim] \gets  0;$\;
			\For{$l = 0$ to $|X_b|$}{
				$x \gets  X_b(l)$\;
				$sumZ \gets 0$\;
				\For{$i = 0$ to $|C|$}{
					$i^* \gets argmax_{k}$ \{teNN$_k(x,r_k,A_{t,k}, A_{c,k})$ s.t. $y_k = C_i$\} \;
					$z_{i^*}\gets $teNN$_{i^*}(x,r_{i^*},A_{t,r_{i^*}}, A_{c,{i^*}})$\; 
					$sumZ \gets sumZ + z_{i^*}$\;
					$\nabla r_{i^*},  \nabla A_{t,i^*}, \nabla A_{c,i^*} \gets$ $Grads(z_{i^*})$\; 
				}
			    \For{$i = 0$ to $|C|$}{	
			    	$o_{i^*}\gets \frac{z_{i^*}}{sumZ}$\; 
			    	\If{$y_{i^*} = y_b(l)$}{
			    		loss $\gets$ loss $- np.log(o_{i^*})$\;
			    		$Gr(i^*) \gets Gr(i^*) - \nabla r_{i^*}*(1/sumZ -1/z_{i^*})$\;
			    		$GA_{t}(i^*) \gets GA_{t}(i^*) - \nabla A_{t,i^*}*(1/sumZ -1/z_{i^*})$\;
			    		$GA_{c}(i^*) \gets GA_{c}(i^*) - \nabla A_{c,i^*}*(1/sumZ -1/z_{i^*}$\;		
		    		}
		    		\Else{
			    		$Gr(i^*) \gets Gr(i^*) - \nabla r_{i^*}*1/sumZ $\;
						$GA_{t}(i^*) \gets GA_{t}(i^*) - \nabla A_{t,i^*}*1/sumZ$\;
						$GA_{c}(i^*) \gets GA_{c}(i^*) - \nabla A_{c,i^*}*1/sumZ$\;			
		    		}
				}
			}
			\For{$k = 0$ to $N_c$}{
				$norm\_factor \gets ||Gr(k)||+||GA_t(k)||+||GA_c(k)||+\epsilon)/f_k$\;
				$R_{k}  \gets r_{k} + \eta Gr(k)/norm\_factor$\;
				$A_{t,k} \gets A_{t,k} + \eta GA_t(k)/norm\_factor$\;
				$A_{c,k} \gets A_{c,k} + \eta GA_c(k)/norm\_factor$\;
			}	
		}
	    \If{No Progress}{ \Comment{No accuracy gain on train data or loss reduction over a given number of epochs}
		$\eta \gets \eta/1.05$\;
	    }
        $nepoch \gets nepoch + 1$\;
	}
\label{alg:training_teNN}
\end{algorithm}

\begin{algorithm}
	\scriptsize 
	\caption{Averaging a set of time series by means of a gradient descent on the KDTW kernel.}\label{alg:TSA}
	\KwData{$X$, $L$, $dim$, $\nu_0$, $max\_epoch$}
	\KwResult{An estimate of the centroid of $X$}
	$nepoch \gets 0$\;
	$M \gets medoid_of(X)$\;
	\While{$nepoch < max\_epoch$}{
		$Gr[L,dim] \gets  0;$\;
		\For{$x \in X$}{
			$G_r  \gets G_r + \nabla KDTW(x,M,\nu_0)$ \Comment{see Eq.\ref{eq:dr} and Eq.\ref{eq:dr_tilde}}
		}
		$M \gets M	 + \eta G_r(k)/(||G_r(k)||)$\;
		$nepoch \gets nepoch + 1$\;
	}
	\label{alg:centroids}
\end{algorithm}

Algorithm~\ref{alg:training_teNN} presents the implementation of a stochastic gradient descent that seeks to minimize the CCE loss on a teNN architecture, namely a set of $N_c$ teNN cells $\{($teNN$_u$, $y_u, f_u$), $u=1\cdots N_c\}$. It takes as arguments a set of labeled time series $(X, y)$, the max number of epochs, the batch size. The main meta parameters are
\begin{enumerate}
	\item $\nu_0$, the bandwidth of the local kernel which is used to initialize the attention matrices $A_t$.
	\item $\alpha_0$, used to initialize the activation matrices $A_c$.
	\item $n_r$ the number of references used to represent each category ($N_r = n_r \dot N_c$).
	\item the relaxation coefficient $\eta >0$.
	\item $\lambda_1 \ge 0$ which weights the sparsity penalty of the activation matrices ($A_c$).
	\item $\lambda_2 \ge 0$ which weights the sparsity penalty of the attention matrices ($A_t$).
\end{enumerate}

The initial references $\{R_u\}_{u=1,\cdots,N_r}$ are determined using algorithm \ref{alg:centroids} which averages a set of time series thanks to a gradient descent on the KDTW kernel, exploiting Eq.\ref{eq:dr} and Eq.\ref{eq:dr_tilde}. If multiple references per category are required, spectral clustering is first applied with the required number of clusters, then the algorithm ~\ref{alg:centroids} is run on each discovered subset of time series.

\section{Validation}
\label{sec:validation}

To validate the teNN architecture and its training algorithm, we propose here after some experiments carried out on three datasets, two are univariate (BME, ECG200) and one is multivariate (ERing). These datasets are selected from the Time Series Classification (TSC) website \footnote{\url{https://timeseriesclassification.com/d}}.

The meta parameters used for these experiments are given in the following table:

\begin{table}[H]
	\centering
	\caption{Selected values for the teNN meta parameters}
	\label{tab:meta_parameters}
	\begin{tabular}{|l|c|l|}
		\toprule
		Parameter & value & comment\\
		\midrule
		$\lambda_t$ & $1e-3$& sparsity penalty for the attention matrices\\
		$\lambda_a$ & $1e-3$& sparsity penalty for the activation matrices\\
		$\eta $ & $1e-1$ & relaxation coefficient\\
		batch\_size & $64$ &\\
		$\nu_0$ & $1e-3$& initialization of the attention matrices $A_t$ \\
		$\alpha_0$ & $1.0$& initialization of the activation matrices $A_c$ \\
		$n_r$ & 1 & the number of reference time series per category.\\
		\bottomrule
	\end{tabular}
\end{table}

The following subsections present for each of these three data sets the contents of matrices $A_t$, $A_c$ and $R$, once the learning procedure has converged. We then evaluate the degree of parsimony of the matrices $A_t$ and $A_c$. We finally carry out an ablation study to estimate the impact of each of these parameter matrices on the accuracy metric.

\subsection{The BME dataset}
\begin{figure}[H]
	\centering
	\includegraphics[scale=0.3]{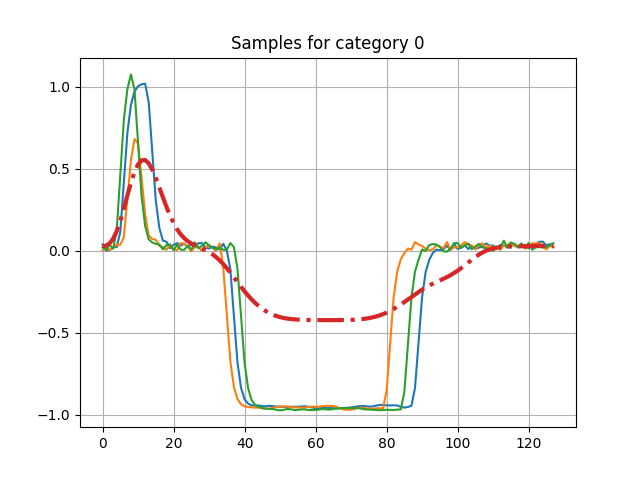}
	\includegraphics[scale=0.3]{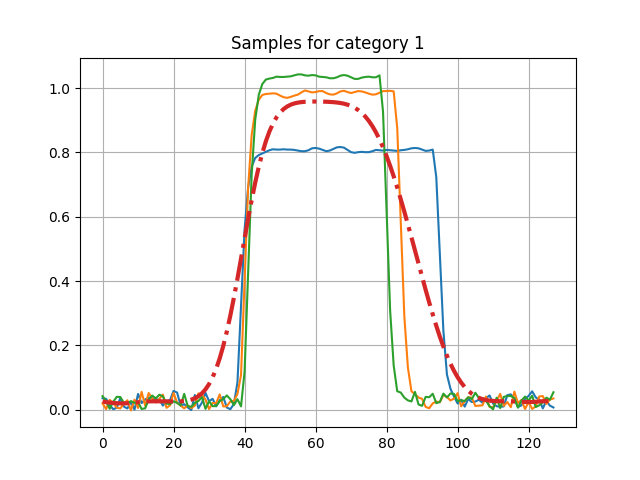}
	\includegraphics[scale=0.3]{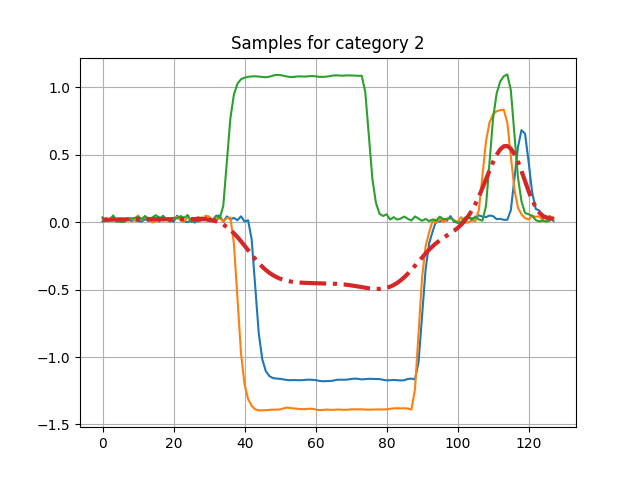}
	\caption{Samples of the BME dataset (3 samples per category are presented). The time elastic centroid used to initialize the references of the teNN cells are represented in red dotted lines.}
	\label{fig:BME-dataset}
\end{figure}

BME (Begin-Middle-End) is a synthetic univariate dataset provided by Laboratoire d'informatique de Grenoble(LIG), at Université Grenoble Alpes. It contains with three classes as shown in Fig.\ref{fig:BME-dataset}: 
\begin{enumerate}
	\item Category 1 series (Begin), are characterized by the presence of a small positive peak arising at the initial period.
	\item Category 2 series (Middle) are characterized by the absence of any peak at the beginning or are the end of the series.
	\item Category 3 series (End) are characterized by the presence of a positive peak arising at the
	 final period.
\end{enumerate}

All series, whatever their category, are constituted by a central plateau. The central plateau may be positive or negative. Hence, the discriminant part is the presence or absence of a positive peak, located at the beginning or at the end of the series.

\begin{figure}[H]
	\centering
	\includegraphics[scale=0.4]{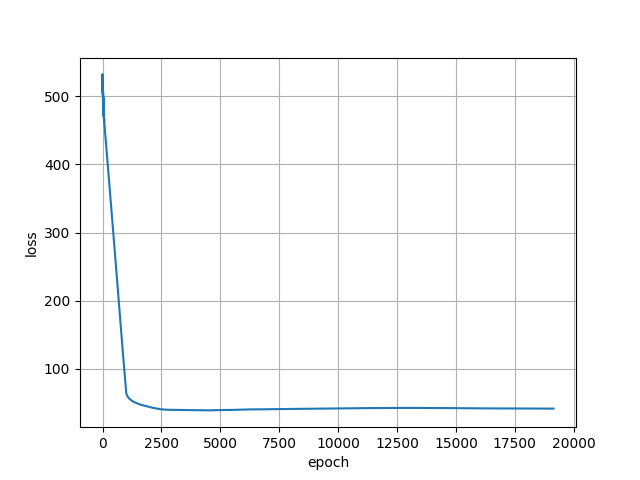}
	\caption{Loss(epoch) for the BME dataset.}
	\label{fig:BME-loss}
\end{figure}	

Fig.\ref{fig:BME-loss} shows the loss function as a function of the epoch for the BME dataset. The convergence appears regular and smooth although a large number of iteration ($>1000$) is required to reach a minimum.

\begin{figure}[H]
	\centering
	\includegraphics[scale=0.3]{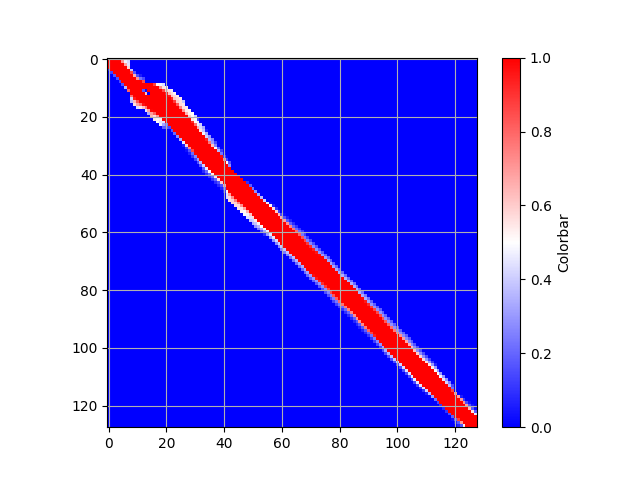}
	\includegraphics[scale=0.3]{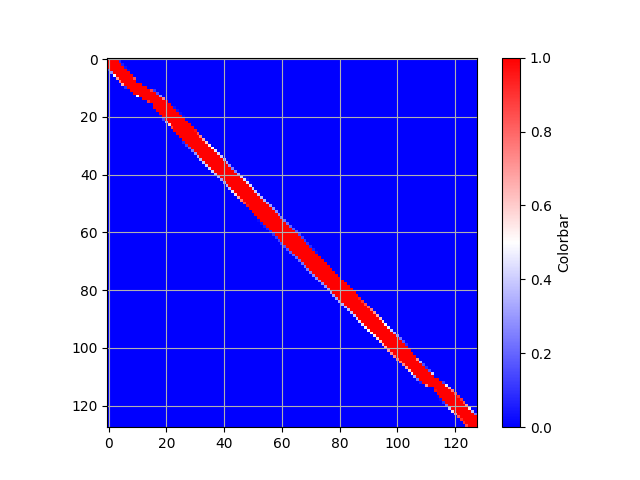}
	\includegraphics[scale=0.3]{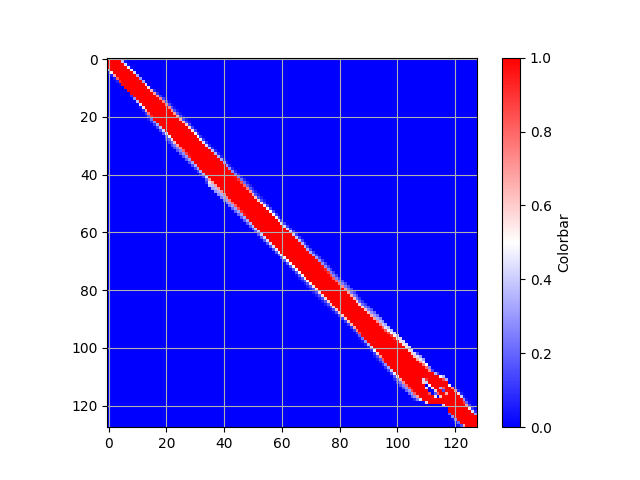}
	\caption{Activation matrices for the BME dataset. From left to right, activation for categories 1 (Begin), 2 (Middle) and 3 (End). }
	\label{fig:BME-activation}
\end{figure}	

The activation matrices $A_c$, one for each of the three categories, are presented as colored encoded images in Fig.\ref{fig:BME-activation}. Red color corresponds to a maximum level of activation ($A_c(i,j) = 1$), while blue color encodes a zero level of activation. On this example, the teNN architecture has been able to learn corridors that limit the search spaces for the alignment paths quite sharply.

\begin{figure}[H]
	\centering
	\includegraphics[scale=0.3]{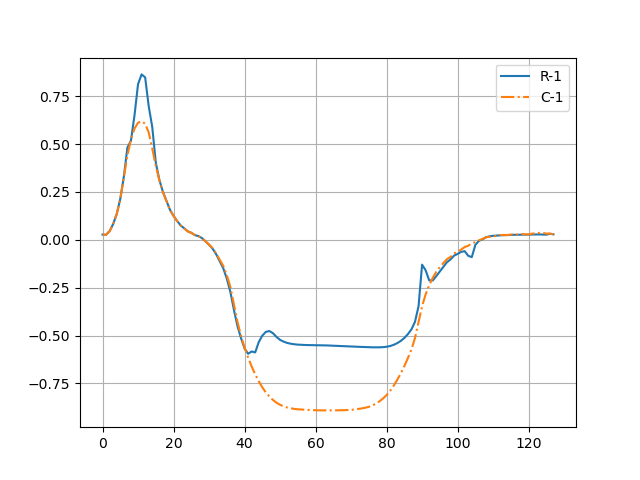}
	\includegraphics[scale=0.3]{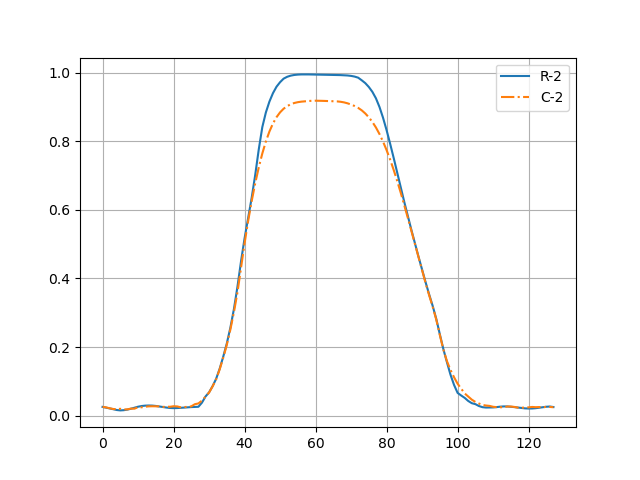}
	\includegraphics[scale=0.3]{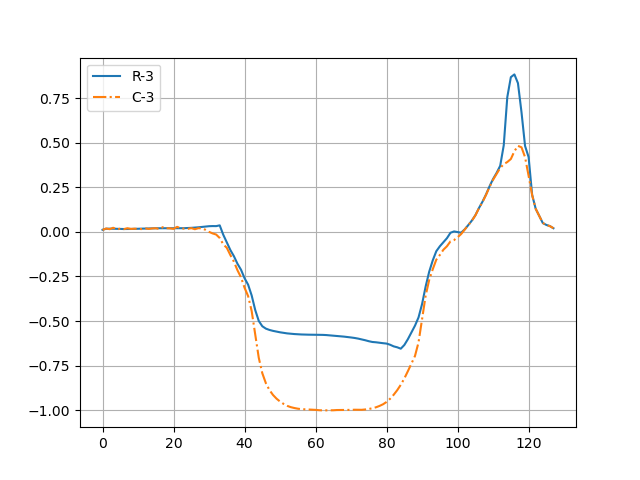}
	\caption{Reference ($\{R_u\}, u \in \{1,2,3\}$) time series for the BME dataset (blue curve, plain line). The centroid of the category used for initializing the teNN cell is presented in orange dotted line. Time is the horizontal axis.}
	\label{fig:BME-references}
\end{figure}	

\begin{figure}[H]
	\centering
	\includegraphics[scale=0.3]{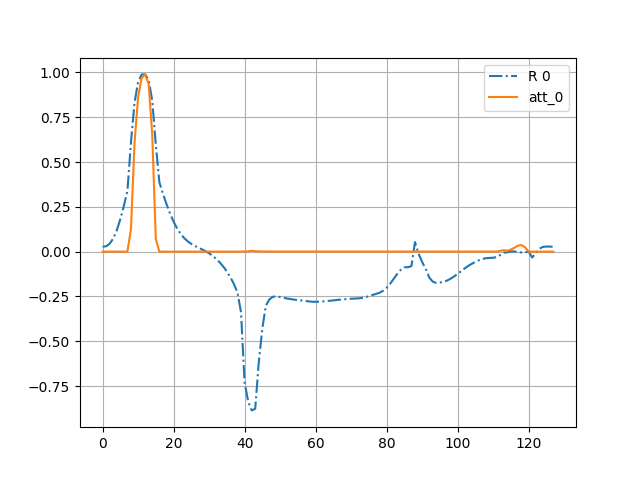}
	\includegraphics[scale=0.3]{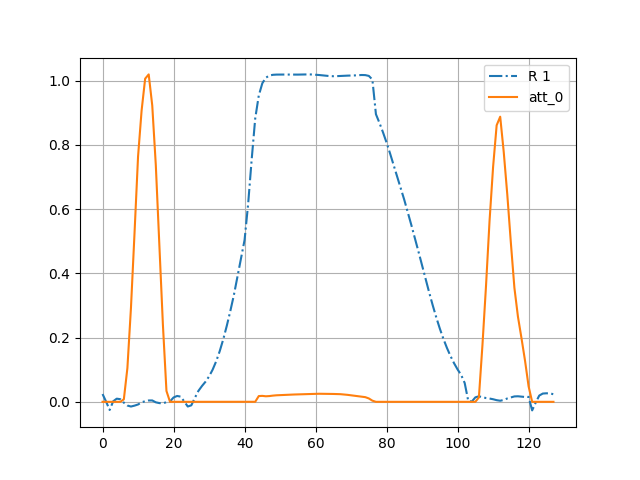}
	\includegraphics[scale=0.3]{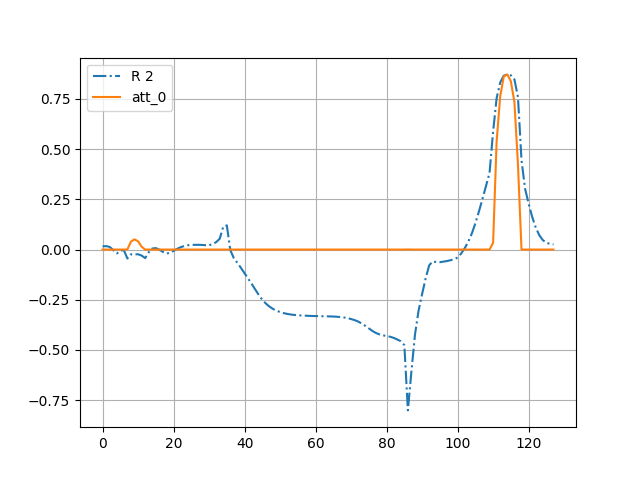}
	\caption{Attention weights ($A_t(i)$)for the BME dataset (orange curve, plain line). The corresponding reference vector $R_u$ is presented in blue dotted line. Time is the horizontal axis.}
	\label{fig:BME-attentions}
\end{figure}	

Since the BME dataset contains univariate time series, the attention matrices $A_t$ reduce to attention vectors that are associated to each of the reference vectors $R$. These vectors are shown in Fig. \ref{fig:BME-attentions}.
 
As expected, the attention of the teNN components are focused on the discriminant part of the reference time series, basically at the beginning and the end of the time series to test the presence or absence of a discriminative peak. The centers of the time series are characterized by a quasi-null attention, meaning that the associated subsequences have no weight in the computation of the output of the teNN component.\\

Fig. \ref{fig:BME-references} shows the initial references for each category in dotted lines. The reference obtained once the training is completed is indicated in solid blue lines. For categories $1$ and $3$, the central part of the final references having zero attention weight is becoming closer to 0. Here, optimizing simultaneously on the references $R$ and on the attention matrices $A_{t}$ is somehow redundant: it sets a null attention on the central plateau and move the reference towards zero, the average value between positive and negative plateaux.

For this example, we could have proposed two reference vectors for categories 1 and 2, to account for positive or negative plateaux in the central part of the series. However, as shown in Fig.~\ref{fig:BME-attentions}, attention weights offer a more parsimonious solution, allowing the use of a single reference per category. 

\subsection{The ECG200 dataset}

\begin{figure}[H]
	\centering
	\includegraphics[scale=1.5]{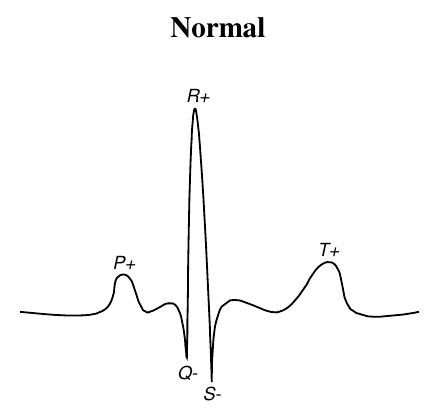}
	\includegraphics[scale=1.5]{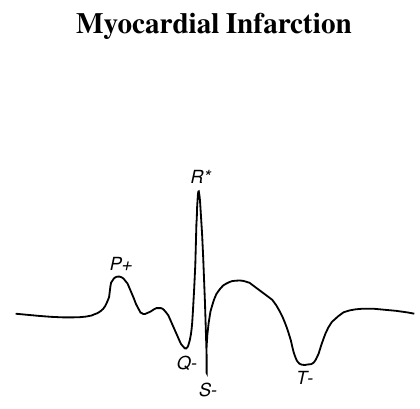}
	\caption{ECG200 categories, Normal is category 0 and Myocardial Infarction is category 1. Figures are from \citep{ECG200_data}}
	\label{fig:ECG200}
\end{figure}

\begin{figure}[H]
	\centering
	\includegraphics[scale=0.3]{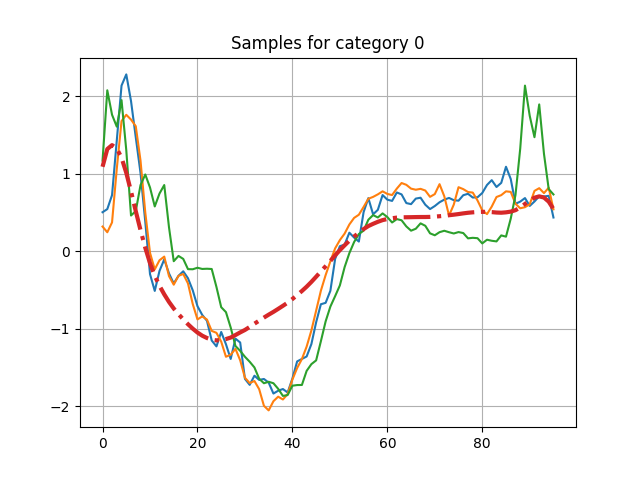}
	\includegraphics[scale=0.3]{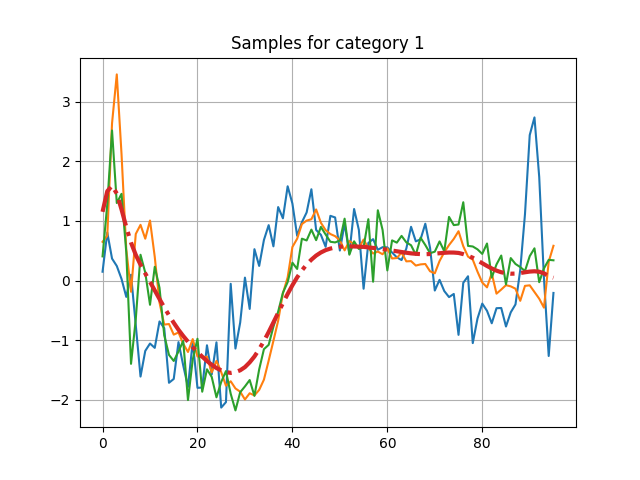}
	\caption{Samples of the ECG200 dataset (3 samples per category are presented). The time elastic centroid used to initialize the references of the teNN cells are represented in red dotted lines.}
	\label{fig:ECG200-dataset}
\end{figure}

For this second dataset~\citep{ECG200_data}, each univariate series is a subsequence of an electrocardiogram corresponding to a single heartbeat. The two classes gather respectively normal heart rhythm time series (category 0) and a myocardial infarction (category 1) time series (typical shapes are given in Fig.\ref{fig:ECG200} from \citep{ECG200_data}). Few time series extracted randomly from the train dataset for each category are presented in Fig.\ref{fig:ECG200-dataset} along with the estimate centroid time series presented in red dash-dot line.

\begin{figure}[H]
	\centering
	\includegraphics[scale=0.4]{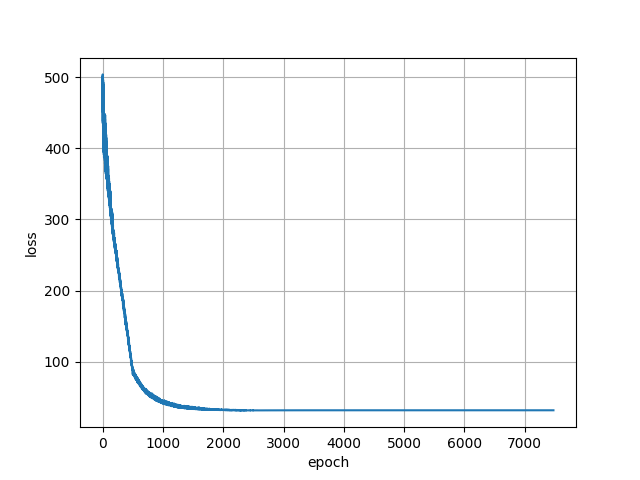}
	\caption{Loss(epoch) for the ECG200 dataset.}
	\label{fig:ECG200-loss}
\end{figure}

Fig.\ref{fig:ECG200-loss} shows the loss function as a function of the epoch for the ECG200 dataset. Here again the convergence appears regular and smooth. The slope of the curve is steep, although a large number of iteration is required before reaching a minimum.

\begin{figure}[H]
	\centering
	\includegraphics[scale=0.3]{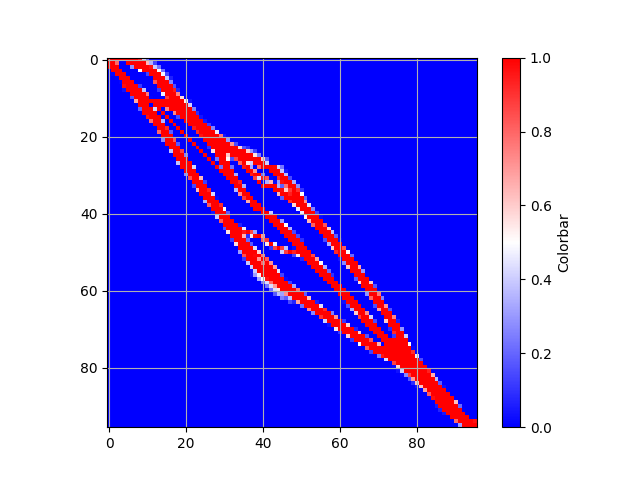}
	\includegraphics[scale=0.3]{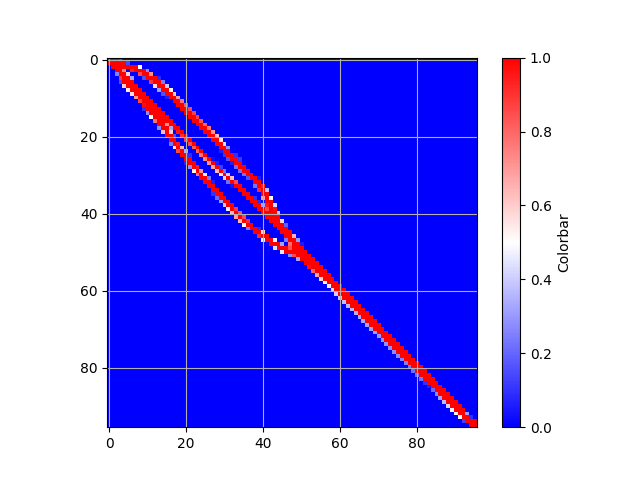}
	\caption{Activation matrices for the ECG200 dataset. From left to right, activation for category $0$ and $1$. }
	\label{fig:ECG200-activation}
\end{figure}	

The activation matrices $A_c$, one for each of the two categories, are presented as colored encoded images in Fig.\ref{fig:ECG200-activation}. Red color corresponds to a maximum level of activation, while blue color encodes a zero level of activation and white color corresponds to in-between activation levels ($A_c(i) \approx .5$). On this example, we show that the teNN architecture has been able to learn complex corridors with 'holes' that forbid the passing of alignment paths. Here again, the search spaces for the alignment paths is reduced quite sharply.

\begin{figure}[H]
	\centering
	\includegraphics[scale=0.3]{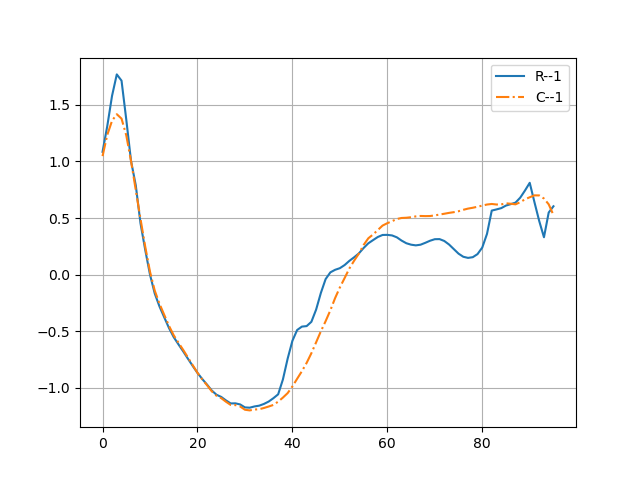}
	\includegraphics[scale=0.3]{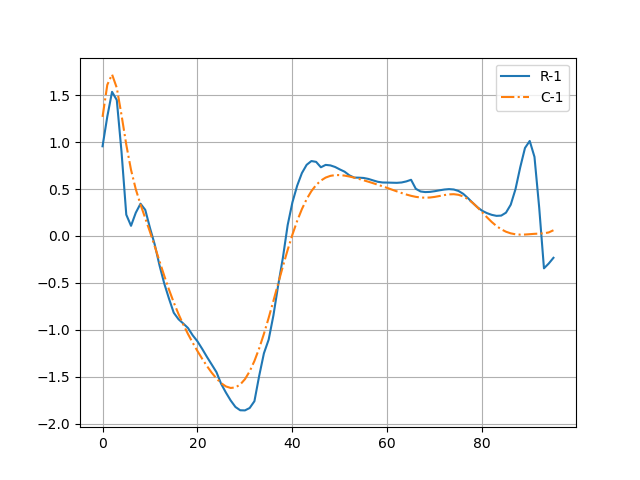}
	\caption{$\{R_u\}$ time series for the ECG200 dataset (blue curve, plain line). The centroid of the category used to initialize the teNN cell is presented in orange dotted line. Time is the horizontal axis.}
	\label{fig:ECG200-R}
\end{figure}

\begin{figure}[H]
	\centering
	\includegraphics[scale=0.3]{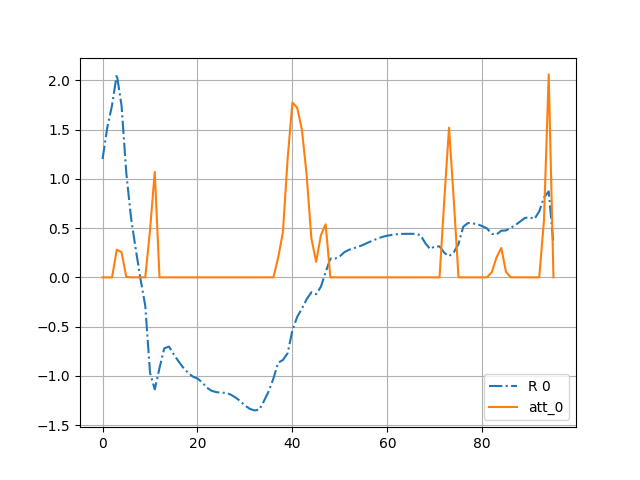}
	\includegraphics[scale=0.3]{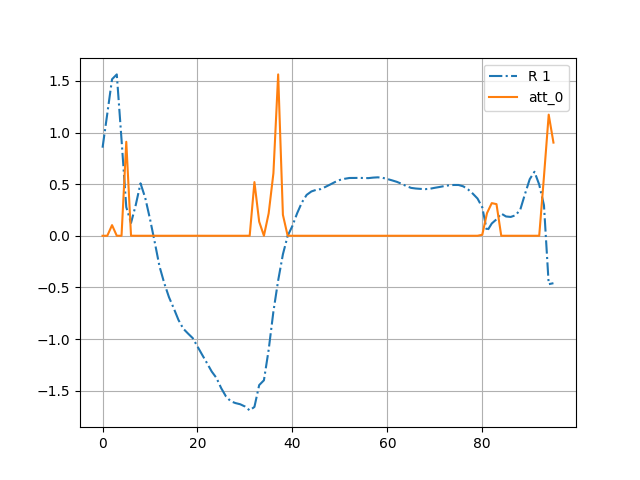}
	\caption{Attention weights for the ECG200 dataset. Left, category 'Normal' (0), right, category Abnormal (1)}
	\label{fig:ECG200-attentions}
\end{figure}	

The ECG200 dataset is composed with univariate time series, hence the attention matrices $\{A_t\}$ reduce again to attention vectors that are associated to each of the reference vectors $R$ shown in Fig.\ref{fig:ECG200-R}. The attention vectors are shown in Fig. \ref{fig:ECG200-attentions} in plain orange curves, while the reference vectors are presented in blue dash-dot curves. The interpretation is not obvious and required some medical expertise, but one can see that the attention vectors focus mostly on the shape of the big negative valley and on the shape of the subsequent plateau.

%
%
%
%
%

\subsection{The ERing dataset}
\begin{figure}
	\centering
	\includegraphics[scale=0.3]{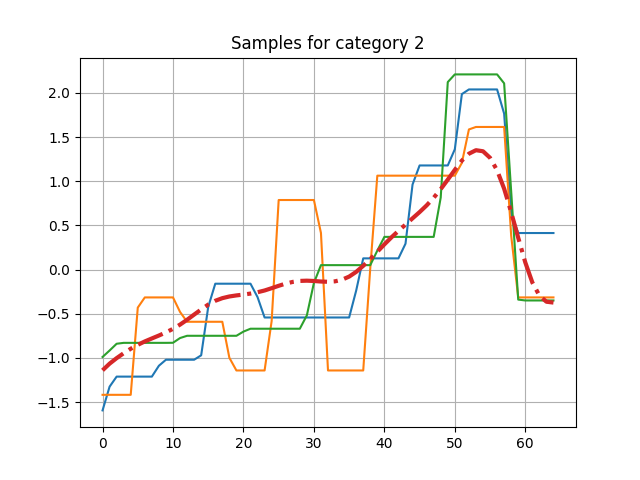}
	\includegraphics[scale=0.3]{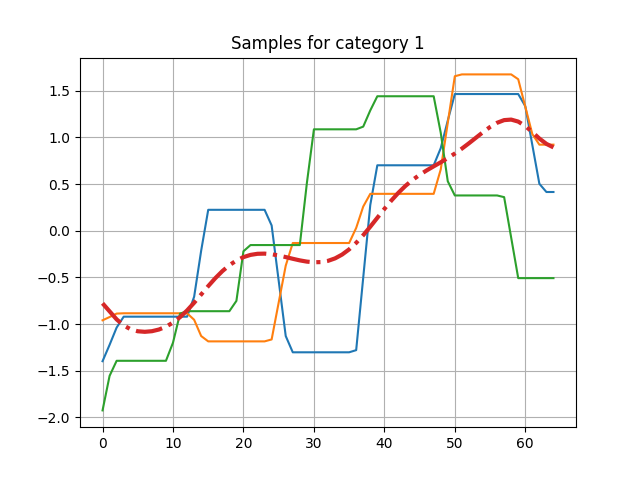}
	\includegraphics[scale=0.3]{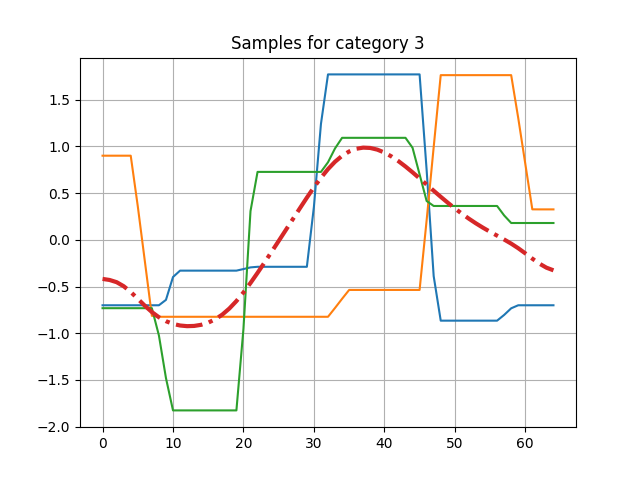}
	\includegraphics[scale=0.3]{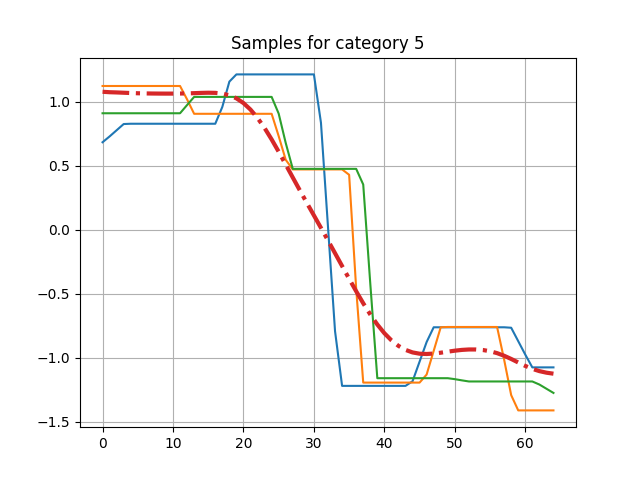}
    \includegraphics[scale=0.3]{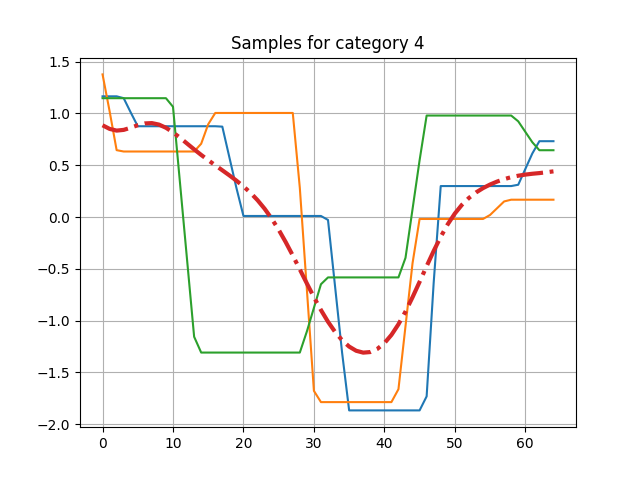}
    \includegraphics[scale=0.3]{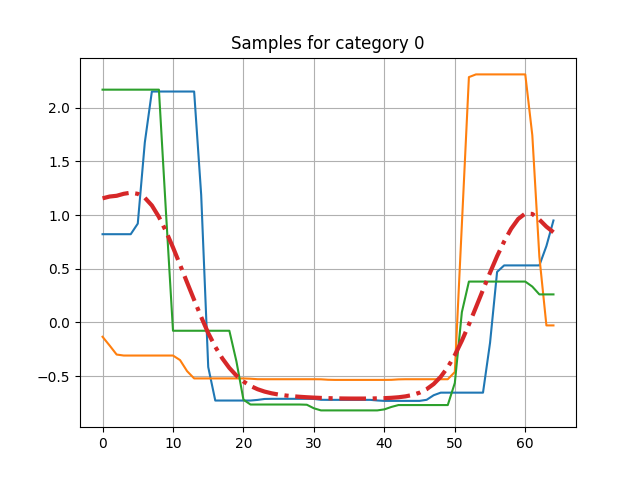}
	\caption{Samples of the ERing dataset. The time elastic centroid used to initialize the references of the teNN cells are represented in red dotted lines.}
	\label{fig:ERing-dataset}
\end{figure}

The ERing dataset \citep{Ering_data} is composed with multivariate times series in 4 dimensions ($d=4$) that characterize captured hand and finger gestures belonging to $6$ categories. These data have been captured using a finger ring and an electromagnetic sensor. Fig.\ref{fig:ERing-dataset} shows few samples for the 6 categories.

\begin{figure}[H]
	\centering
	\includegraphics[scale=0.4]{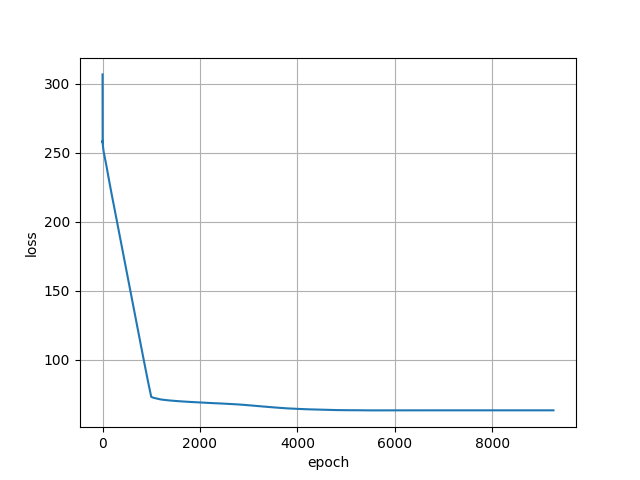}
	\caption{Loss as a function of the epoch for the ERing dataset.}
	\label{fig:ERing-loss}
\end{figure}	

Similarly to the previous examples, for the ERing multivariate data, the loss function presented in Fig.\ref{fig:ERing-loss} shows a smooth and regular convergence of the training process.

\begin{figure}[H]
	\centering
	\includegraphics[scale=0.3]{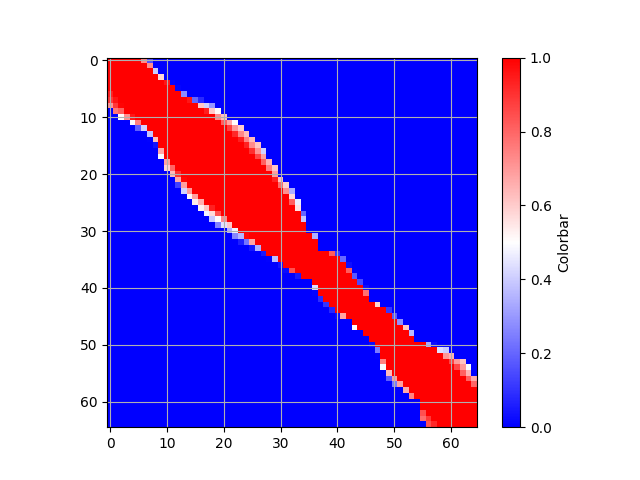}
	\includegraphics[scale=0.3]{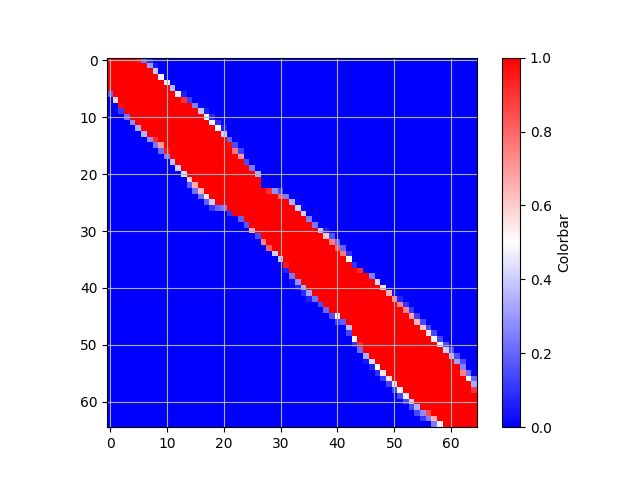}
	\includegraphics[scale=0.3]{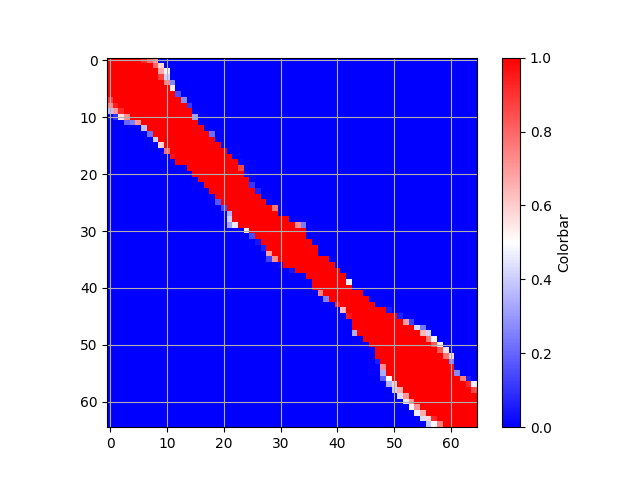}
	\includegraphics[scale=0.3]{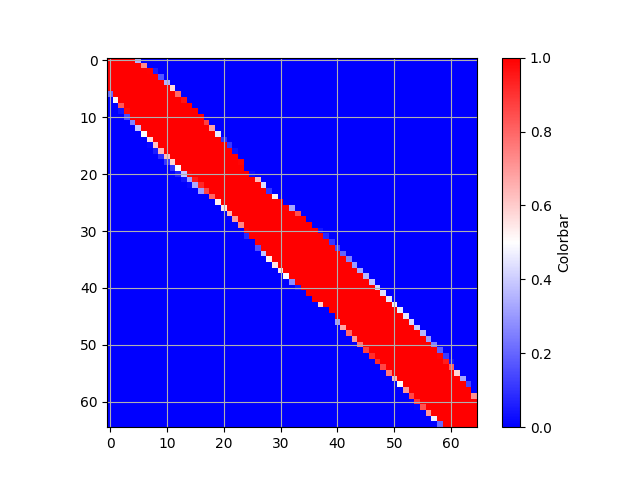}
	\includegraphics[scale=0.3]{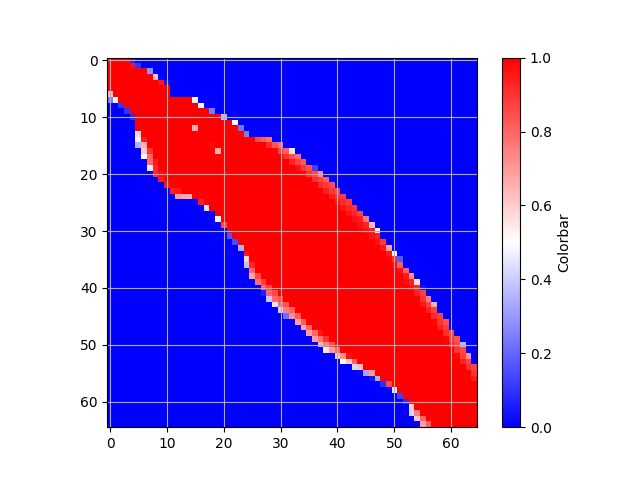}
	\includegraphics[scale=0.3]{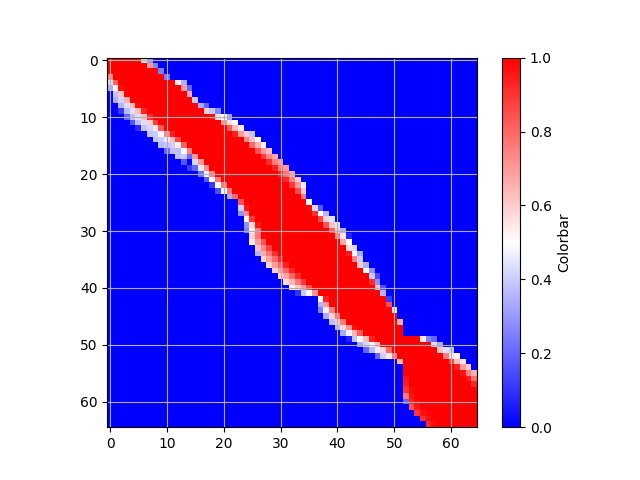}
	\caption{Activation matrices for the ERing dataset.}
	\label{fig:ERing-activation}
\end{figure}	

The activation matrices given in Fig.\ref{fig:ERing-activation} are very contrasted. Either the neurons are activated (red pixels) or deactivated (blue pixels). On this example, the teNN cells were able to significantly reduce the alignment search spaces, as shown by the narrow corridors.

\begin{figure}[H]
	\centering
	\includegraphics[scale=0.8]{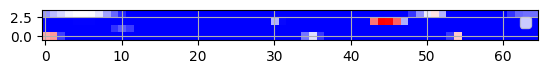}
	\includegraphics[scale=0.8]{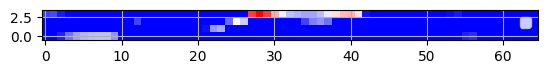}
	\includegraphics[scale=0.8]{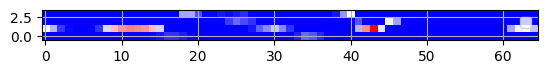}
	\includegraphics[scale=0.8]{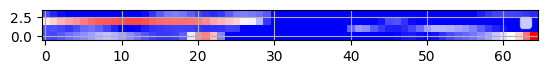}
	\includegraphics[scale=0.8]{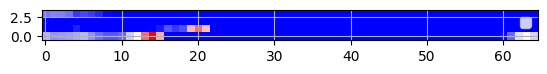}
	\includegraphics[scale=0.8]{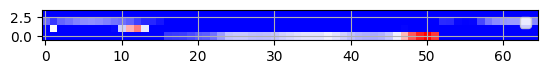}
	\caption{Attention weights for the ERing dataset. Red pixels stand for high attention, blue ones for low attention. White pixels identify in between level of attention.}
	\label{fig:ERing-attentions}
\end{figure}

As the time series are multivariate, the attention matrices do not reduce to vectors. They are shown in Fig.\ref{fig:ERing-attentions}. From this example, we see that, after training, the teNN architecture was able to focus its attention on a spatio-temporal basis, basically, at any given time, attention is generally focused on a small subset of dimensions. The attention matrix is sparse which was expected using a L1 penalty (meta parameter $\lambda_2$ in Eq.\ref{eq:loss}).

\begin{figure}[H]
	\centering
	\includegraphics[scale=0.3]{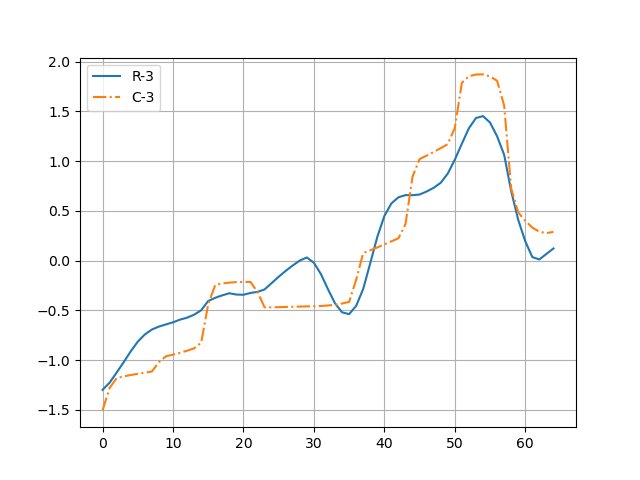}
	\includegraphics[scale=0.3]{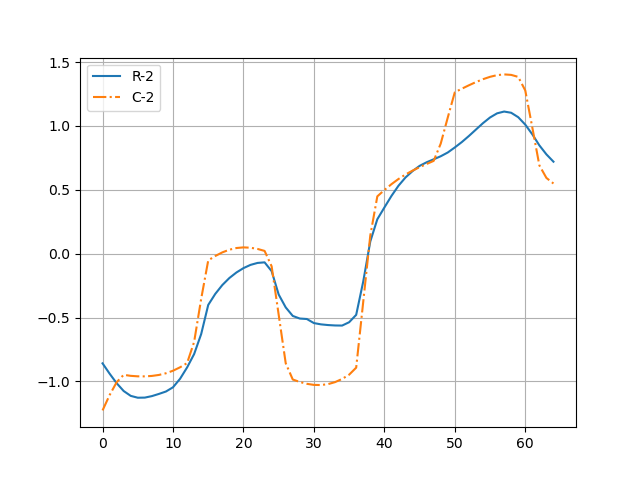}
	\includegraphics[scale=0.3]{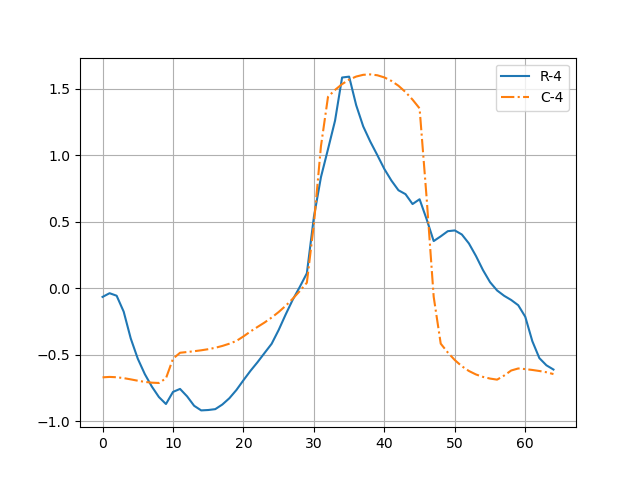}
	\includegraphics[scale=0.3]{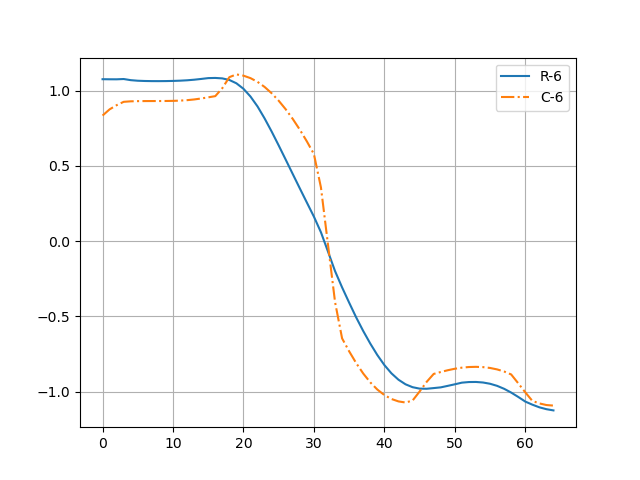}
	\includegraphics[scale=0.3]{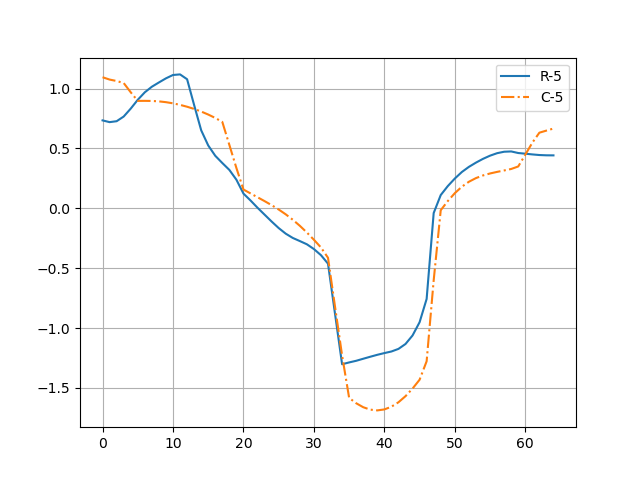}
	\includegraphics[scale=0.3]{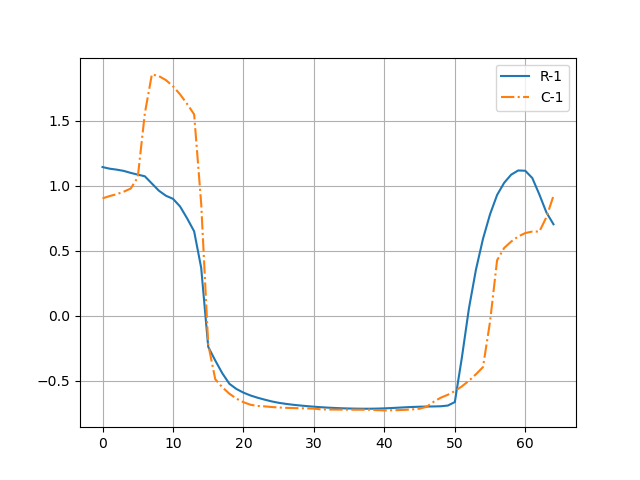}
	\caption{Reference ($\{R_i\}$) time series for the ERing dataset.}
	\label{fig:ERing-references}
\end{figure}	

Fig.\ref{fig:ERing-references} shows the reference time series after learning. For this example, they are staying close to the initial centroids although some details of the shapes have been erased or added. Without a precise knowledge of the motion  capture process, we cannot interpret further this example.

\subsection{How parsimonious is the teNN model?}
Beyond limiting the number of class references, we need to estimate the sparsity of the teNN components, namely the $A_c$ and $A_t$ matrices. The sparsity of any matrix can be evaluated as the \% of zero elements it contains. 
The sparsity of the matrix $A_c$ estimates the size of the useful corridor in which the alignment paths could be confined. For the matrix $A_t$, it indicates the area of the reference time series $R$ on which the network focuses its attention. Zero values correspond to unimportant local sample alignments. Although the teNN architecture is based on the KDTW global alignment kernel, $A_t$ sparsity allows extracting local segments with high attention interconnected via "don't care" segments. Somehow, this provides teNN with local alignment capability.

Note that there is no reason for the matrix $R$, which represents a reference time series, to be sparse.
\begin{table}[H]
	\centering
	\caption{Sparsity (in \%) of the $A_c$ and $A_t$ matrices of the teNN architecture}
	\label{tab:sparsity_study}
	\begin{tabular}{|l|l|l|l|}
		\toprule
		& BME & ECG200 & ERing \\
		\midrule
		$A_c$ & 90.0 &87.7& 75.7\\
		$A_t$ & 62.5 & 68.8 & 69.0 \\
		\bottomrule
	\end{tabular}
\end{table}

Table \ref{tab:sparsity_study} shows the sparsity of the matrices $A_c$ and $A_t$ once the teNN architecture has been trained on the BME, ECG200 and ERing datasets with meta-parameters $\lambda_{t}$ and $\lambda_{c}$ set to the value $1e^{-2}$. For the synthetic BME dataset the sparsity of matrix $A_c$ reaches 90\%, while it is a bit less for the ECG200 and the ERing datasets ($\approx$ 87\% and 75\% respectively). The sparsity of the $A_t$ matrix is also significant ($\approx$ 62\% tà 69\%) for the three datasets.

\subsection{Ablation study}

To evaluate the importance of the three main components ($A_c$, $A_t$ and $R$ matrices) which compose the teNN architecture, we have carried out an ablation study and compared the classification accuracies obtained on the three previous datasets. 

\begin{table}[H]
	\centering
	\caption{Accuracies (in \%) obtained when none, one or several components of the teNN architecture are not optimized. First value corresponds to the last minimum training error configuration, while the second value corresponds to the minimum loss configuration. Best accuracies are in bold characters.}
	\label{tab:ablation_study}
	\begin{tabular}{|l|l|l|l|}
		\toprule
		& BME & ECG200 & ERing \\
		\midrule
		$R$ only &72/71.33 &84/82 & 86.67/84.44\\
		$A_c$ only &74.67/74.67 & 78/78& 86.30/86.30\\
		$A_t$ only &\textbf{100}/85.33 & 87/87& 93.70/91.48\\
		$R$ and $A_c$ & 70.67/70.67 & 90/91 &86.67/85.19 \\
		$R$ and $A_t$ &100/99.33 & 84/87& 93.70/86.30\\
		$A_c$ and $A_t$ & 100/83.33&86/88 &94.44/89.26 \\
		\textbf{Full model} &\textbf{100/100} &\textbf{92/92} & \textbf{95.18/95.18} \\
		\midrule
		1NN-DTW & 89.33 & 80.00 & 93.33\\
		1NN-KDTW & 98.66 & 83 & 92.59\\
		\bottomrule
	\end{tabular}
\end{table}

Table \ref{tab:ablation_study} presents the accuracies obtained for the fully optimized teNN architecture (last raw), and for configurations in which one or to of the three components ($A_c$, $A_t$ and $R$) are not optimized. When $A_t$ is not optimized, the attention matrix is initialized by a constant $\nu$ value selected such as minimizing the training error of a one near neighbor classifier (using a leave one out procedure).

Clearly, this study shows that the joint optimization of the three components leads to the best accuracies. Apparently, the component having the most impact on the drop in accuracy is the attention matrix $A_t$. The $R$ and $A_c$ component have a significant impact on the ECG200 task, but not so much on the BME and ERing tasks. The $A_c$ component, as already mentioned, primarily serves to balance precision and parsimony. Not optimizing it, as expected, does not have a catastrophic impact on accuracy.

As a baseline, we added in the table \ref{tab:ablation_study} the accuracies of the first DTW and KDTW near neighbor classifiers. We observe that the full teNN architecture achieves better classification accuracies even though it uses a single reference time series $R$ to represent each category.

\subsection{Wrapping-up the first results}
At the light of these few previous examples, we have shown that with a single reference time series per category, the teNN classifier outperforms the 1-NN classifier based on KDTW or DTW, which leads to a significant speedup. Furthermore, we have established that:
\begin{enumerate}
	\item the stochastic gradient descent is effective to train the teNN architecture, as the loss functions decreases rapidly in general,
	\item the teNN architecture is able to learn sparse attention matrices that reduce the computational cost of the teNN cells and can be interpretable if expertise is available. 
	\item the teNN is able to learn part of its interconnection by deactivating unnecessary neurons. This makes it possible to learn complex alignment corridors and opens prospects for optimizing the implementation of teNN, notably by reducing its memory requirements and computational complexity.
\end{enumerate}

\section{Experimentation}

\label{sec:experimentation}
The preliminary tests we have performed on univariate time series classification shows that teNN achieves comparable results to those presented in \citep{Kewei2021}. In this section, we focus on multivariate time series classification tasks, which are more difficult and prone to overfitting.  
We therefore compare the teNN architecture on these tasks with some state-of-the-art methods, while following the extensive empirical study that was designed and carried out in \citep{Baldan2021}.

\normalsize
	
\subsection{Datasets}

The datasets made up of multivariate time series on which the evaluation is carried out are those used in this previous study \citep{Baldan2021}. These are 30 datasets from various fields (economy, health, biology, energy, industry, etc.), freely accessible on the UEA archive\footnote{\url{https://timeseriesclassification.com/}}.

The characteristics of these datasets (size, length of series, dimension) are given in table \ref{tab:UEA-dataset-info} directly provided in \citep{Baldan2021}. 

Since the current implementation of the teNN architecture requires a fixed length for time series, when the series in a dataset are of variable size, we reduce them all to the length of the longest series by zero-padding.

\begin{table}[!h]
	\centering
	\caption{Datasets information from the UEA repository}
	\label{tab:UEA-dataset-info}
	\resizebox{.9\textwidth}{!}{
		\begin{tabular}{|l|l|l|l|l|l|}
			\toprule
			Dataset                   & Train & Test  & Length & Dims & Class \\
			\midrule
			ArticularyWordRecognition & 275   & 300   & 144    & 9    & 25    \\
			AtrialFibrillation        & 15    & 15    & 640    & 2    & 3     \\
			BasicMotions              & 40    & 40    & 100    & 6    & 4     \\
			CharacterTrajectories     & 1422  & 1436  & 60-182 & 3    & 20    \\
			Cricket                   & 108   & 72    & 1197   & 6    & 12    \\
			DuckDuckGeese             & 50    & 50    & 270    & 1345 & 5     \\
			EigenWorms                & 128   & 131   & 17984  & 6    & 5     \\
			Epilepsy                  & 137   & 138   & 206    & 3    & 4     \\
			EthanolConcentration      & 261   & 263   & 1751   & 3    & 4     \\
			ERing                     & 30    & 270   & 65     & 4    & 6     \\
			FaceDetection             & 5890  & 3524  & 62     & 144  & 2     \\
			FingerMovements           & 316   & 100   & 50     & 28   & 2     \\
			HandMovementDirection     & 160   & 74    & 400    & 10   & 4     \\
			Handwriting               & 150   & 850   & 152    & 3    & 26    \\
			Heartbeat                 & 204   & 205   & 405    & 61   & 2     \\
			InsectWingbeat            & 25000 & 25000 & 2-22   & 200  & 10    \\
			JapaneseVowels            & 270   & 370   & 7-29   & 12   & 9     \\
			Libras                    & 180   & 180   & 45     & 2    & 15    \\
			LSST                      & 2459  & 2466  & 36     & 6    & 14    \\
			MotorImagery              & 278   & 100   & 3000   & 64   & 2     \\
			NATOPS                    & 180   & 180   & 51     & 24   & 6     \\
			PenDigits                 & 7494  & 3498  & 8      & 2    & 10    \\
			PEMS-SF                   & 267   & 173   & 144    & 963  & 7     \\
			PhonemeSpectra            & 3315  & 3353  & 217    & 11   & 39    \\
			RacketSports              & 151   & 152   & 30     & 6    & 4     \\
			SelfRegulationSCP1        & 268   & 293   & 896    & 6    & 2     \\
			SelfRegulationSCP2        & 200   & 180   & 1152   & 7    & 2     \\
			SpokenArabicDigits        & 6599  & 2199  & 4-93   & 13   & 10    \\
			StandWalkJump             & 12    & 15    & 2500   & 4    & 3     \\
			UWaveGestureLibrary       & 120   & 320   & 315    & 3    & 8    \\
			\bottomrule
		\end{tabular}
	}
\end{table}

\subsection{Compared methods}

In \citep{Baldan2021}, the authors developed a feature based approach called Compexity Measures and Features for Multivariate Time series (CMFM) to improve the interpretability of classification results. More precisely, time series are represented with 41 descriptive features (in particlar curvature, linearity, Shannon entropy, skewness, trend, etc.). Several models of classification are evaluated, namely C5.0 with boosting (CMFM-C5.0B), Random Forest (CMFM-RF), Support vector machine (CMFM-SVM) and 1-Nearest Neighbors with Euclidean distance (CMFM-1NN-ED). 

In addition, they included in their experimental study the following SOTA models :
\begin{itemize}
	\item 1-Nearest Neighbors with Euclidean distance (1NN-ED).
	\item 1-Nearest Neighbors with DTW distance using multidimensional points (1NN-DTW-D), with or without normalization \citep{ShokoohiYekta2016}.
	\item 1-Nearest Neighborsbased on the sum of 1 dimensional DTW distances (1NN-DTW-I), with or without normalization \citep{ShokoohiYekta2016}.
	\item Multivariate LSTM with fully Convolutional Networks (MLSTM-FCN) \citep{Karim2018} with the settings specified by the authors.
	\item Word ExtrAction for time SEries cLassification plus Multivariate Unsupervised Symbols and dErivatives (WEASEL + MUSE) \citep{schafer2018} with the settings specified by their authors.
	\item Local Cascade Ensemble for Multivariate data classification (LCEM) \citep{Fauvel2020}, optimized hyper-parameters for each dataset.
	\item Random Forest for Multivariate (RFM) algorithm, from the sklearn library, applied to the transformation proposed in the 	LCEM paper \citep{Fauvel2020}.
	\item Extreme Gradient Boosting for multivariate (XGBM), Extreme Gradient Boosting algorithm, from the xgboost library,applied to the transformation proposed in the LCEM paper \citep{Fauvel2020}.
\end{itemize}

All the reported classification results from these methods presented in Table \ref{tab:results} are from \citep{Fauvel2020} and \citep{Baldan2021}. We use them for comparison purposes, to position the teNN architecture within the state of the art in the field.

\subsection{Heuristic selection for the teNN meta parameters}
The convergence of the training process depends strongly on the choice of three meta-parameters: $\eta$, the relaxation parameter, $\nu0$, the initialization of the attention matrices and $n_r$ the number of reference time series per category.

The heuristic we have adopted consists in selecting the highest values for $\eta$ and $\nu0$ and the lowest value for $n_r$ so that the learning procedure converges.

\begin{table}[H]
	\centering
	\caption{Selected values for the teNN meta parameters}
	\label{tab:meta_parameters_exp}
	\begin{tabular}{|l|c|l|}
		\toprule
		Parameter & value & comment\\
		\midrule
		$\lambda_t$ & $1e-6$& sparsity penalty for the attention matrices\\
		$\lambda_a$ & $1e-6$& sparsity penalty for the activation matrices\\
		$\eta $ & $1e-1$ & relaxation coefficient\\
		batch\_size & size of dataset &\\
		$\alpha_0$ & $1.0$ & initialization of the activation matrices $A_c$ \\
		\bottomrule
	\end{tabular}
\end{table}

For the other meta parameters, we have retained the values listed in Table \ref{tab:meta_parameters_exp}.

\subsection{Results}
\label{Results}

\begin{table}[]
	\centering
	\caption{Accuracy results on the UEA repository datasets: accuracy (\%), average accuracy, median, average rank, and Win/Loss/Tie Ratio. Best results are in bold.}
	\label{tab:results}
	\begin{sideways}
		\resizebox{1.33\textwidth}{!}{
			\begin{tabular}{|l|l|l|l|l|l|l|l|l|l|l|l|l|l|l|l|l|l|}
				\toprule
				Datasets                  &  \makecell{CMFM+\\C5.0B} & \makecell{CMFM+\\RF}                    & \makecell{CMFM+\\SVM} & \makecell{CMFM+\\1NN-ED} & LCEM                          & XGBM                          & RFM                            & \makecell{MLSTM\\-\\FCN}                     & \makecell{WEASEL\\+\\MUSE}                   & \makecell{ED-\\1NN} & \makecell{DTW-\\1NN-I}                     & \makecell{DTW-\\1NN-D}                     & \makecell{ED-\\1NN\\(norm)} & \makecell{DTW-\\1NN-I\\(norm)}               & \makecell{DTW-\\1NN-D\\(norm)}    & teNN-lm & teNN-ml            \\
				\midrule
				ArticularyWordRecognition & 91         & 99                              & 97.7        & 98.3           & \textbf{99.3} & 99                              & 99                              & 98.6                            & \textbf{99.3} & 97     & 98                              & 98.7                            & 97            & 98                              & 98.7      &  98.3 & 98.3                      \\
				AtrialFibrilation         & 20         & 20                              & 26.7        & 13.3           & \textbf{46.7} & 40                              & 33.3                            & 20                              & 26.7                            & 26.7   & 26.7                            & 20                              & 26.7          & 26.7                            & 22                       & 33.3&33.3       \\
				BasicMotions              & 85         & 97.5                            & 92.5        & 95             & \textbf{100} & \textbf{100} & \textbf{100} & \textbf{100} & \textbf{100} & 67.5   & \textbf{100} & 97.5                            & 67.6          & \textbf{100} & 97.5         & 100 & 100                  \\
				CharacterTrajectories     & 95.3       & 97.1                            & 95.9        & 90.7           & 97.9                            & 98.3                            & 98.5                            & \textbf{99.3} & 99                              & 96.4   & 96.9                            & 99                              & 96.4          & 96.9                            & 98.9                       & 99.2 & 99.2     \\
				Cricket                   & 86.1       & 97.2                            & 95.8        & 97.2           & 98.6                            & 97.2                            & 98.6                            & 98.6                            & 98.6                            & 94.4   & 98.6                            & \textbf{100} & 94.4          & 98.6                            & \textbf{100} & 99.2 & 99.2\\
				DuckDuckGeese             & 42         & 52                              & 44          & 40             & 37.5                            & 40                              & 40                              & \textbf{67.5} & 57.5                            & 27.5   & 55                              & 60                              & 27.5          & 55                              & 60                             & 42.0& 42.0\\
				EigenWorms                & 81.7       & 88.5                            & 84          & 79.4           & 52.7                            & 55                              & \textbf{100} & 80.9                            & 89                              & 55     & 60.3                            & 61.8                            & 54.9          & 61.8                            & NA                             & 0& 0\\
				Epilepsy                  & 88.4       & \textbf{100} & 97.8        & 95.7           & 98.6                            & 97.8                            & 98.6                            & 96.4                            & 99.3                            & 66.7   & 97.8                            & 96.4                            & 66.6          & 97.8                            & 96.4                      & 93.5 &93.5      \\
				Ering                     & 80.7       & \textbf{93} & 92.6        & 90.4           & 20                              & 13.3                            & 13.3                            & 13.3                            & 13.3                            & 13.3   & 13.3                            & 13.3                            & 13.3          & 13.3                            & 13.3                         & 96.3& 96.3  \\
				EthanolConcentration      & 35         & 33.5                            & 32.7        & 30.4           & 37.2                            & 42.2                            & \textbf{43.3} & 27.4                            & 31.6                            & 29.3   & 30.4                            & 32.3                            & 29.3          & 30.4                            & 32.3                         & 31.2&24.3  \\
				FaceDetection             & 54         & 54.8                            & 57.9        & 50.5           & 61.4                            & \textbf{62.9} & 61.4                            & 55.5                            & 54.5                            & 51.9   & 51.3                            & 52.9                            & 51.9          & 52.9                            & NA                          & 65.8 & 65.8    \\
				FingerMovements           & 44         & 52                              & 46          & 53             & 59                              & 53                              & 56                              & \textbf{61} & 54                              & 55     & 52                              & 53                              & 55            & 52                              & 53                           & 63.0 & 63.0   \\
				HandMovementDirection     & 33.8       & 28.4                            & 32.4        & 18.9           & \textbf{64.9} & 54.1                            & 50                              & 37.8                            & 37.8                            & 27.9   & 30.6                            & 23.1                            & 27.8          & 30.6                            & 23.1                         & 45.91 & 45.9   \\
				Handwriting               & 16.5       & 28.2                            & 18.4        & 24.9           & 28.7                            & 26.7                            & 26.7                            & 54.7                            & 53.1                            & 37.1   & 50.9                            & \textbf{60.7} & 20            & 31.6                            & 28.6                       & 40.7 & 40.7     \\
				Heartbeat                 & 74.1       & 76.6                            & 73.2        & 62             & 76.1                            & 69.3                            & \textbf{80} & 71.4                            & 72.7                            & 62     & 65.9                            & 71.7                            & 61.9          & 65.8                            & 71.7                          & 72.7 & 72.7  \\
				InsectWingbeat            & NA         & \textbf{64.0} & 10          & 25.8           & 22.8                            & 23.7                            & 22.4                            & 10.5                            & 12.8                            & 11.5   & 12.8                            & NA                              & NA            & NA                              & NA                          & 55.4 & 55.4    \\
				JapaneseVowels            & 82.4       & 87.6                            & 76.5        & 72.2           & 97.8                            & 96.8                            & 97                              & \textbf{99.2} & 97.8                            & 92.4   & 95.9                            & 94.9                            & 92.4          & 95.9                            & 94.9                         & 97.3 & 97.3   \\
				Libras                    & 83.9       & 86.7                            & 83.3        & 82.8           & 77.2                            & 76.7                            & 78.3                            & \textbf{92.2} & 89.4                            & 83.3   & 89.4                            & 87.2                            & 83.3          & 89.4                            & 87                        & 86.7 & 86.7      \\
				LSST                      & 63.1       & \textbf{65.2} & 64.8        & 50             & \textbf{65.2} & 63.3                            & 61.2                            & 64.6                            & 62.8                            & 45.6   & 57.5                            & 55.1                            & 45.6          & 57.5                            & 55.1                        & 43.4 &  43.4  \\
				MotorImagery              & 49         & 51                              & 50          & 44             & \textbf{60.0} & 46                              & 55                              & 53                              & 50                              & 51     & 39                              & 50                              & 51            & 50                              & NA                         & 58.0 &  58.0   \\
				NATOPS                    & 81.7       & 81.7                            & 75          & 73.9           & 91.6                            & 90                              & 91.1                            & \textbf{96.1} & 88.3                            & 85     & 85                              & 88.3                            & 85            & 85                              & 88.3                       & 93.3 &  93.3   \\
				PenDigits                 & 93.3       & 95.1                            & 95.9        & 94.4           & 97.7                            & 95.1                            & 95.1                            & \textbf{98.7} & 96.9                            & 97.3   & 93.9                            & 97.7                            & 97.3          & 93.9                            & 97.7                        & 96.2 & 96.2    \\
				PEMSF                     & 96.5       & \textbf{100} & 69.4        & 77.5           & 94.2                            & 98.3                            & 98.3                            & 65.3                            & 70.5                            & 73.4   & 71.1                            & 70.5                            & 73.4          & 71.1                            & NA                       & 74.0 &72.1       \\
				PhonemeSpectra                   & 22.4       & 28.7                            & 25          & 15.8           & \textbf{28.8} & 18.7                            & 22.2                            & 27.5                            & 19                              & 10.4   & 15.1                            & 15.1                            & 10.4          & 15.1                            & 15.1                      & 862 & 8.44     \\
				RacketSports              & 72.4       & 80.9                            & 80.9        & 71.1           & \textbf{94.1} & 92.8                            & 92.1                            & 88.2                            & 91.4                            & 86.4   & 84.2                            & 80.3                            & 86.8          & 84.2                            & 80.3                      & 86.8 & 86.8      \\
				SelfRegulationSCP1        & 81.2       & 81.2                            & 79.2        & 70.3           & 83.9                            & 82.9                            & 82.6                            & \textbf{86.7} & 74.4                            & 77.1   & 76.5                            & 77.5                            & 77.1          & 76.5                            & 77.5                      & 86.3 & 86.0      \\
				SelfRegulationSCP2        & 53.9       & 41.7                            & 46.1        & 50             & \textbf{55} & 48.3                            & 47.8                            & 52.2                            & 52.2                            & 48.3   & 53.3                            & 53.9                            & 48.3          & 53.3                            & 53.9                      & 56.1 & 56.1      \\
				SpokenArabicDigits        & 93.9       & 96.9                            & 97.5        & 92.6           & 97.3                            & 97                              & 96.8                            & \textbf{99.4} & 98.2                            & 96.7   & 96                              & 96.3                            & 96.7          & 95.9                            & 96.3                      & 98.7 & 98.7      \\
				StandWalkJump             & 26.7       & 33.3                            & 20          & 13.3           & 40                              & 33.3                            & \textbf{46.7} & \textbf{46.7} & 33.3                            & 20     & 33.3                            & 20                              & 20            & 33.3                            & 20                        & 40.0 & 40.0      \\
				UWaveGestureLibrary       & 64.1       & 77.2                            & 73.8        & 75.3           & 89.7                            & 89.4                            & 90                              & 85.7                            & \textbf{90.3} & 88.1   & 86.9                            & \textbf{90.3} & 88.1          & 86.8                            & \textbf{90.3} & 86.9 & 86.9\\
				\midrule
				\textbf{Average Rank}               & 12       & 8.5           & 11         & 13           & \textbf{5.1}               & 8        & 6.4           & 6.2        & 7         & 12   & 10                             & 8.9                             & 12          & 11                             & 9.4                     &6.2 &   6.4     \\
				\textbf{Win/Loss/Tie Ratio}        & 0/30/0 &  4/26/0 &  0/30/0 &  0/30/0 &  \textbf{7/23/2} &  1/29/1 &  5/25/2 &  10/20/2 &  3/27/3 &  0/30/0 &  1/29/1 &  3/27/2 &  0/30/0 &  1/29/1 &  2/28/2 &  5/25/5 &  5/25/5\\                    
				\bottomrule	
			\end{tabular}
		}
	\end{sideways}
\end{table}

Two teNN models were evaluated. The first, teNN-lm, corresponds to the last minimum learning error strategy. The second, teNN-ml, corresponds to the last minimum loss strategy. The classification accuracies obtained by these two models are shown in the last two columns of the Table \ref{tab:results}.

The last column of the Table \ref{tab:results} shows the average rank achieved by each model. 

\begin{figure}[H]
	\centering
	\includegraphics[scale=0.5]{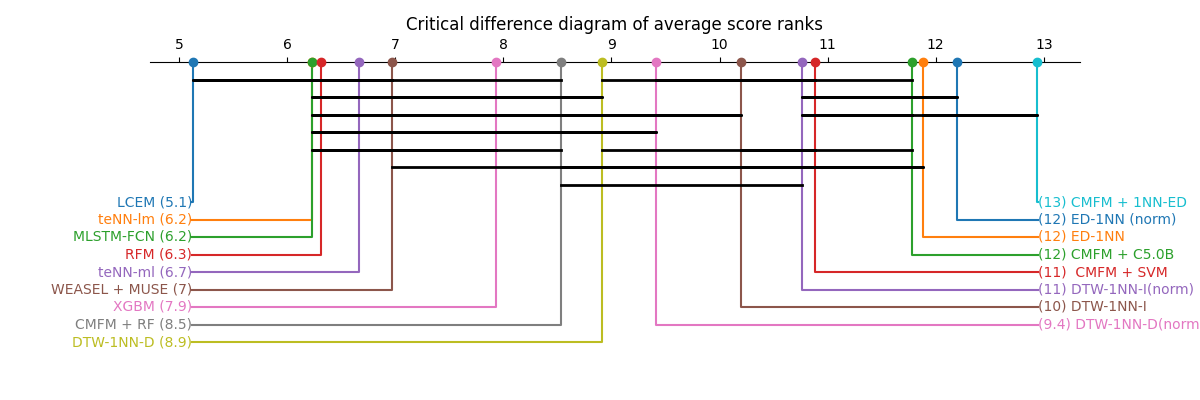}
	\caption{Critical difference diagram}
	\label{fig:CDD}
\end{figure}	

As a post-hoc test, we considered the Wilcoxon signed rank test. The null hypothesis for this test is that the difference between the accuracies of the two compared models is zero. If the p-value is below the chosen significance level (0.05), we reject the null hypothesis and conclude that there is a significant difference between the accuracies of the two models. Otherwise, we consider that there is no significant difference between the two models.

Fig. \ref{fig:CDD} shows the Critical Difference Diagram (CDD) for the set of evaluated models. CDD is a way of visualizing post-hoc test statistics. Firstly, in a block design scenario, the values within each block are ranked, and the average rank across all blocks for each treatment is plotted along the x axis. A crossbar is then drawn over each group of treatments that do not show a statistically significant difference among themselves. Here, the significance level is set to $0.05$ (less than 5\%).\\

Based on this benchmark, we conclude that the teNN architecture has classification performance very similar to that of state-of-the-art models, in particular Random Forest and MLSTM-FCN models. The only model that performs better, although not significantly so according to the Wilcoxon post-hoc analysis, is the LCEM model, which is a meta-model combining Random Forest and XGBoost. The gain/loss/tie ratios are very close between teNN and the best models, LCEM, CMFM+RF and MLSTM-FCN.

Furthermore, the two tested decision strategies, last minimum training error (teNN-lm) and last minimum loss (teNN-ml) lead to the similar classification accuracies, although teNN-lm achieves a slightly better average rank.

\section{Conclusion}

We have introduced and detail in this article an atypical neural network architecture for multivariate time series classification purpose. More precisely, we have proposed a neural network architecture that explicitly incorporates time warping capability. The approach we have develop is rooted into the theory of kernels \citep{Schoenberg38}, which are essentially similarity measures to which corresponds an inner product in the so-called Reproducing Kernel Hilbert Space.

Behind the design of this architecture, our overall objective was threefold: firstly, we were aiming at improving the accuracy of instance based classification approaches that shows quite good performances as far as enough training data is available. Secondly we have seek to reduce the computational complexity inherent to these methods to improve their scalability. In practice, we have tried to find an acceptable balance between these first two criteria. And finally, we have seek to enhance the explainability of the decision provided by this kind of neural architecture.


The novelty of the teNN, compared to classical neural networks is the following.
\begin{enumerate}
	\item The complete network architecture is an assembly of competitive subnetworks called teNN cells. Each teNN cell is associated with three main components: i) an abstract time series, called reference, ii) an activation matrix and iii) an attention matrix.
	\item In a cell, the output of any elementary neuron is the sum of its inputs (at most three) multiplied by the local kernel which evaluates the pairwise correspondence of time series samples (thus, samples of the input time series input are compared to that of the reference time series).
	\item For each neuron, the inverse of the local kernel bandwidth is a parameter ($\nu$) that is learned during training. A high value means high local attention, while a low value means low attention, basically an area in the time series where we don't care about sample comparison. All attention parameters (inverse of bandwidth values) are grouped within a teNN cell into an attention matrix.
	\item each elementary neuron is associated with an activation weight learned during training. Therefore, neurons inactivated after training can be removed to simplify the neuronal architecture. In a way, the network is able to learn a drop-out strategy, hence optimizing its own architecture. All activation weights in a teNN cell are grouped into an activation matrix.
	\item Finally, the reference time series samples are also adapted during training, which provides interpretable discriminative cues in areas where attention is high.
\end{enumerate}

The experiment demonstrates that the stochastic gradient descent implemented to train a teNN is quite effective. To the extent that the selection of some critical meta-parameters is correct, convergence is generally smooth and fast.

While maintaining good accuracy, we show a drastic gain in scalability by first reducing the required number of reference time series, i.e. the number of teNN cells required. Secondly, we demonstrate that, during the training process, the teNN succeeds in reducing the number of neurons required within each cell. Finally, we show that the analysis of the activation and attention matrices as well as the reference time series after training provides relevant information to interpret and explain the classification results.

The comparative study that we have carried out and which concerns around thirty diverse and multivariate datasets shows that the teNN obtains results comparable to those of the state of the art, in particular similar to those of a network mixing LSTM and CNN architectures for example.\\

However, the study of TeNN architecture is still in its infancy. The impact of meta-parameters (initialization of attention matrices, number of reference time series, relaxation parameter, etc.) needs to be studied in detail.

The current implementation of the architecture itself also needs to be optimized to reduce memory requirements, in particular, inactivated neurons have to be effectively pruned to recover memory space. Very long and high dimensional time series will not fit in memory. Segmentation or multiresolution approaches could be considered to handle this issue. Finally, the implementation could be adapted to run on highly parallel computing platforms, i.e. GPUs and FPGAs. 

\section{Bibliography}
\bibliography{biblio.bib}{}

\end{document}